\RequirePackage{fix-cm}
\documentclass[twocolumn]{svjour3}
\smartqed 

\usepackage{graphicx}
\usepackage{comment}
\usepackage{amsmath,amssymb}
\usepackage{color}
\usepackage{tabularx,booktabs}
\newcolumntype{Y}{>{\centering\arraybackslash}X}
\newcolumntype{L}[1]{>{\hsize=#1\hsize\raggedright\arraybackslash}X}%
\newcolumntype{R}[1]{>{\hsize=#1\hsize\raggedleft\arraybackslash}X}%
\newcolumntype{C}[1]{>{\hsize=#1\hsize\centering\arraybackslash}X}%
\usepackage{multirow}
\usepackage{url}
\usepackage{enumerate}
\usepackage{algpseudocode}
\usepackage{algorithm}
\newcommand{\algrule}[1][.2pt]{\par\vskip.5\baselineskip\hrule height #1\par\vskip.5\baselineskip}

\begin{document}

\title{Learning Condition Invariant Features for Retrieval-Based Localization from 1M Images}
\titlerunning{Learning Features for Retrieval-Based Localization}

\author{Janine Thoma$^{1}$\and
Danda Pani Paudel$^{1}$\and
Ajad Chhatkuli$^{1}$\and 
Luc Van Gool$^{1,2}$
}
\authorrunning{J. Thoma et al.}

\institute{%
Janine Thoma\\ \email{jthoma@vision.ee.ethz.ch} \\~\\
Danda Pani Paudel\\ \email{paudel@vision.ee.ethz.ch} \\~\\
Ajad Chhatkuli\\ \email{ajad.chhatkuli@vision.ee.ethz.ch} \\~\\
Luc Van Gool\\ \email{vangool@vision.ee.ethz.ch} \\~\\
$^1$ Computer Vision Lab, ETH Zurich, Switzerland \\
$^2$ VISICS, ESAT/PSI, KU Leuven, Belgium
}

\date{Uploaded: 9 July 2020}

\maketitle

\maketitle
\begin{abstract}
Image features for retrieval-based localization must be invariant to dynamic objects (e.g. cars) as well as seasonal and daytime changes. Such invariances are, up to some extent, learnable with existing methods using triplet-like losses, given a large number of diverse training images. However, due to the high algorithmic training complexity, there exists insufficient comparison between different loss functions on large datasets. In this paper, we train and evaluate several localization methods on three different benchmark datasets, including Oxford RobotCar with over one million images. This large scale evaluation yields valuable insights into the generalizability and performance of retrieval-based localization. Based on our findings, we develop a novel method for learning more accurate and better generalizing localization features. It consists of two main contributions: (i) a feature volume-based loss function, and (ii) hard positive and pairwise negative mining. On the challenging Oxford RobotCar night condition, our method outperforms the well-known triplet loss by 24.4\% in localization accuracy within 5m. 
\end{abstract}

\section{Introduction}
Vision-based localization has the potential to become an effective solution for several important applications ranging from robotics to augmented reality. A high quality, view aware\footnote{Images captured with an intention to localize with a wide view of the surrounding.} image often captures sufficient information to uniquely represent a location. It is therefore not surprising that we humans use vision as primary source of information for localization, navigation, and exploration in our environments.
In the same spirit, vision-based systems have the potential to offer accurate and robust localization when GPS information and GPS-tagged maps are not reliable or entirely unavailable (e.g.\ indoors).

Traditionally, vision-based localization~\cite{xin2019review} is tackled with structure-based methods, such as Structure-from-Motion (SfM)~\cite{Irschara2009,Li2010location,Sattler2017,Sattler2017pami,Zeisl2015,choudhary2012visibility} and Simultaneous Localization and Mapping (SLAM)~\cite{Mouragnon2006,davison2007monoslam,castle2008video,Bresson2017,eade2006scalable}, or with retrieval-based approaches~\cite{krizhevsky2012imagenet,Simonyan2014,Arandjelovic2014,arandjelovic2016netvlad,dusmanu2019d2,magliani2018accurate,sattler2018benchmarking,hermans2017defense}.
Structure based methods usually focus on accurate relative pose estimation, while retrieval-based approaches prioritize absolute re-localization.
In fact, localization by image retrieval (or simply retrieval-based localization) is inexpensive, with simpler mapping and matching techniques, which also scales better to larger spaces~\cite{arandjelovic2016netvlad,Sattler2017,Thoma2019}. Many structure-based approaches also use retrieval for initialization~\cite{Sattler2017}. 

The problem of retrieval-based localization equates to matching one or more query images, taken at some unknown location, to a set of geo-tagged reference images.
Recent developments in learning image features for object and place recognition~\cite{krizhevsky2012imagenet,Simonyan2014,Arandjelovic2014,arandjelovic2016netvlad} have made image retrieval a viable method for localization. Despite their huge potential, retrieval-based localization methods often suffer from matching inaccuracies, especially when the reference and query images are captured under different conditions~\cite{Anoosheh2018} including day-time, weather, season, or dynamic scene changes. Therefore, image features for retrieval-based localization must be invariant to such changes. 
However, in the literature, the task of learning condition invariant features for retrieval-based localization has received little to no attention. Most of the related methods mainly focus on place/landmark recognition
~\cite{arandjelovic2016netvlad,Simonyan2014,Arandjelovic2014,Kim2017}, where the features learned to recognize places are shown to somewhat generalize to retrieval-based localization.

Previous methods also do not pay special attention to invariances (other than relying on the place recognition dataset already capturing them), nor do they use a task specific setup.
In fact, methods for place recognition primarily aim at distinguishing between prominent landmarks---where images do not necessarily have to be geo-tagged\footnote{Images from different locations can observe the same place/landmark.}.
Therefore, place recognition features may not work very well for accurate localization. Furthermore, there is a lack of thorough study on how these methods perform at localization when trained on a diverse localization dataset containing examples of the desired invariances. While the recently proposed benchmark~\cite{sattler2018benchmarking} is a positive step in that direction, very large scale evaluations are still not prioritized in most works.

\begin{figure*}[t]
 \centering
 \includegraphics[width=\textwidth]{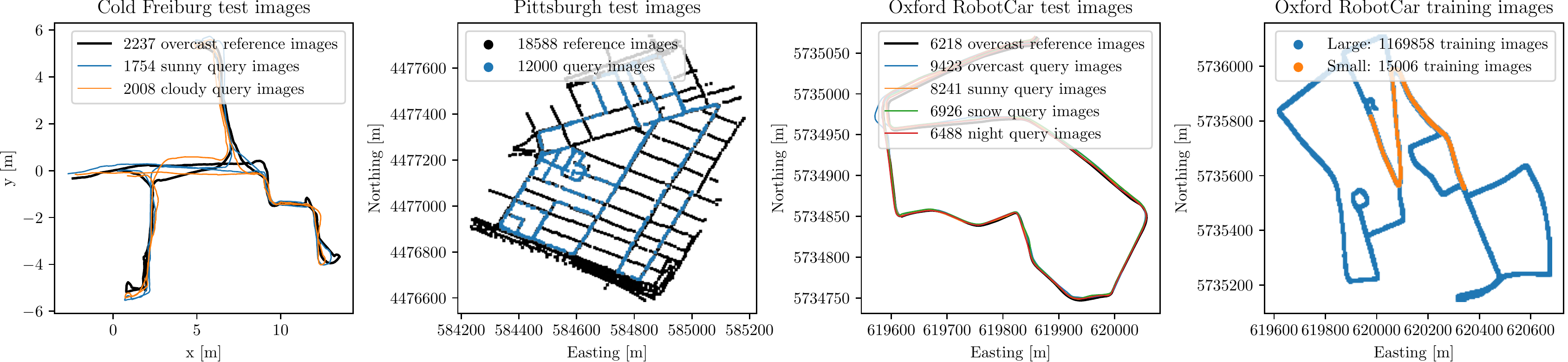}
 \caption{Left to right: test reference and test query images for Cold Freiburg (two query conditions), Pittsburgh, Oxford  RobotCar (four query conditions). Right: training images for smaller (II) and larger (III) Oxford  RobotCar training set.}
\label{fig:test_sets}
\vspace{-0mm}
\end{figure*}

The problem of retrieval-based localization involves learning a feature embedding from a sparse set of geo-tagged images, also known as reference/landmark images. The learned image features together with the corresponding geo-locations serve as a map, in which the query images are localized. In this setup, the location of the query image is inferred by finding the nearest neighbour, of its feature, in the map. The simplicity of localization by finding the nearest neighbour makes the retrieval-based localization efficient and scalable. This however, demands an appropriate feature embedding invariant to the factors not involved in defining the location. In this work, we aim at learning such embeddings using a corpus of geo-tagged images captured under diverse conditions of desired invariances. During this process, we benefit from large scale datasets, examples shown in Figure~\ref{fig:test_sets}, together with the power of discriminative and descriptive feature learning of Convolutional Neural Networks~\cite{krizhevsky2012imagenet,Simonyan2014,schroff2015facenet}.

\begin{table*}[t]
\begin{centering}
\resizebox{0.8\textwidth}{!}{%
\begin{tabular}{@{}clcr@{}}
\toprule
 & Training regions   & Number of Images  & Conditions \\
\midrule 
I & Cold Freiburg   & 29,237  & cloudy, sunny \\ 
II & Oxford  RobotCar (small) & 15,006 &   night, snow, overcast, sun\\
III & Oxford  RobotCar (large) & 1,169,858  & night, snow, overcast, sun, clouds, rain, dusk \\
IV & Pittsburgh    & 30,000  & unspecified \\
V & ImageNet    & \textgreater{}10M & unspecified\\
\bottomrule
\end{tabular}%
}
\caption{Number of training images and conditions for experimental setups I-V. \label{tab:training_size}}%
\end{centering}
\end{table*}

A promising direction for learning feature embeddings, from large scale geo-tagged images, includes methods that make use of triplet/quadruplet loss functions (or their variants)~\cite{schroff2015facenet,weinberger2006distance,chen2017beyond,Yu_2019_ICCV,wang2019ranked}, where hard examples are mined for faster learning and better representation. 
One large-scale dataset that is particularly suitable for this task is Oxford  RobotCar~\cite{maddern20171}. In spite of having access to large-scale datasets of the desired kind (please, refer Table~\ref{tab:training_size}), there is a lack of thorough study on how the existing methods perform on them, when it comes to localization. This lack can primarily be attributed to the algorithmic complexity of training strategies of over $O(n^3)$ with respect to the number of images. This calls for an efficient method that allows us to meaningfully learn localization features from large scale datasets. In fact, improving methods that rely on hard example mining involves two major aspects to be considered: (i) an efficient sampling strategy for mining, and (ii) a powerful objective function to minimize. 

This paper addresses the problem of learning accurate and generalizing localization features from large scale datasets. Within the framework of hard example mining for metric learning, we propose two major contributions concerned with sampling strategy and objective function, suitable for the task at hand. Furthermore, we also provide valuable practical and theoretical insights regarding the factors affecting localization to facilitate the discussion about invariance, by comparing the proposed method against various existing ones. To summarize, the major contributions of this paper are threefold:
\begin{enumerate}
 \item \textbf{Mining for invariance:} Hard positive and pairwise negative mining.
 \item \textbf{Feature volume-based objective:} Efficient for large scale datasets.
 \item \textbf{Experiments/insights with $>1$M images:} Factors affecting localization.
\end{enumerate}

In the following, we will show how the proposed mining, especially the hard positives, helps us, and is meaningful, to learn invariant features within the considered experimental setup. For each anchor image, several negative and positive images are mined. The objective for learning is then defined using these images by approximating the volume covered by their features. In Section~\ref{sec:experiments}, we show that the volume-based objective function is a better choice over the commonly used distance-based alternatives. 
Using the proposed method, we learn features on three benchmark datasets, including Oxford  RobotCar with over one million images. Our experiments demonstrate the superiority of our features in terms of both invariance and accuracy, with an improvement of $24.4\%$ within 5m on the challenging Oxford RobotCar night condition when compared to triplet loss~\cite{arandjelovic2016netvlad}. In Section~\ref{sec:discussion}, we provide an insightful discussion, for the first time to the best of our knowledge, based on the proposed and other methods, about the considerations that need to be made while learning features for localization.

\section{Related work}
There is a large volume of work which tackles image-based localization. In this section we briefly summarize some relevant works on image retrieval-based localization, interested readers can refer to \cite{zheng2017sift,piasco2018survey} for more details. In retrieval-based localization, differences largely lie on the feature learning aspects; once discriminative features have been learned, matching is a relatively simple task. Features useful for recognition as well as localization are generally learned in the paradigm of metric learning. The triplet loss~\cite{schroff2015facenet,weinberger2006distance} learns image features that are useful for recognition. They tackle manifold learning on images by jointly minimizing feature distances between positive samples and maximizing the feature distances between negative samples. The method and its variants have been proven to be highly useful for image retrieval~\cite{hermans2017defense}. Inspired by its success, several other methods have improved the training loss by providing hard negative mining~\cite{vysotska2016lazy} and simultaneously pushing away negative pairs from the positive ones~\cite{chen2017beyond}, known as the quadruplet loss. In order to improve on the quadruplet loss, very recently \cite{Yu_2019_ICCV} proposed an n-tuplet loss for metric learning. Triplet and quadruplet losses use mining strategies in order to find the best samples to learn from. A recent work~\cite{Duan_2019_CVPR} proposes a method to improve mining of triplets composing of hard negatives for training. In order to accelerate triplet loss by avoiding such sampling schemes, \cite{Qian_2019_ICCV} proposes a smooth version of the triplet loss that shows good performance for certain tasks. Specifically for the objective of retrieval-based localization, \cite{thoma2020geometrically} propose adding the geometric distance between locations as a smooth regularizer for metric learning. \cite{Zhong_2019_ICCV} addresses the issue of adversarial attacks on training with triplet loss. Metric learning is a highly sought problem and several other improvements or variations have been proposed. A non-exhaustive list includes the use of multiple pairwise distances lifted to higher dimension \cite{oh2016deep}, adversarial learning \cite{duan2018deep,xu2019deep}, angular loss \cite{wang2017deep}, adaptation to the multi-class problem \cite{kim2019deep}, and learning of local features~\cite{dusmanu2019d2,revaud2019r2d2,mishchuk2017working,luo2019contextdesc,noh2017large}.

A slightly different approach to improving features involves learning to pool relevant features while discarding others (also reducing the feature dimensionality). NetVLAD~\cite{arandjelovic2016netvlad} does exactly so by pooling lower dimensional descriptive features for localization on top of VGGNet~\cite{Simonyan2014} using the triplet loss and has been very influential in retrieval-based localization. \cite{Kim2017} trains a network to further discard NetVLAD~\cite{arandjelovic2016netvlad} features that are irrelevant for localization. \cite{tolias2015particular} proposes R-mac, also a pooling strategy, that was later trained end-to-end in \cite{Gordo2017} using the Siamese architecture. \cite{magliani2018accurate} improves upon R-mac~\cite{tolias2015particular} specifically for landmark recognition.
In the same spirit, \cite{seddati2017towards} uses R-mac in order to train CNN feature predictors for image retrieval.

In other efforts, \cite{sattler2018benchmarking} has proposed evaluation schemes for 6dof pose estimation with a variety of datasets.
Another line of research concerns directly addressing seasonal or day-night variations either by using 3D point clouds~\cite{angelina2018pointnetvlad} or by domain transfer~\cite{Anoosheh2018}.
Although, in most use cases, Structure-based localization methods benefit from retrieval, \cite{shen2018matchable} shows that a 3D surface reconstruction can actually benefit retrieval-based localization learning using overlapping images.
It is worth noting that, apart from feature learning, image retrieval can also benefit from works on improving feature matching~\cite{Philbin2007,Stumm2015}, as well as map image summarization~\cite{Thoma2019}, and exploiting temporal closeness of images~\cite{milford2012seqslam,milford2015sequence}.

\section{Learning from 1M Images}

\begin{figure*}
\centering
\def\svgwidth{\textwidth}
 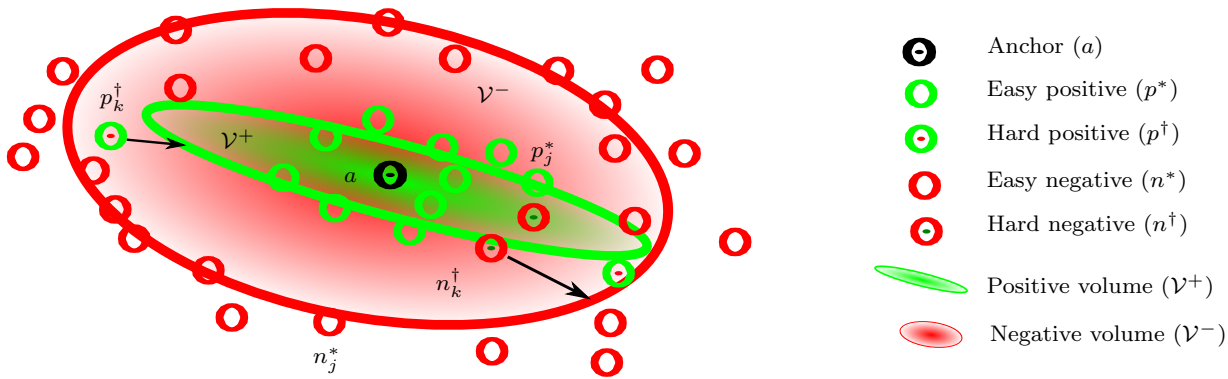
\caption{Positive and negative features, together with their respective volume, visualized for an anchor point. We push hard positive towards and hard negatives away from the anchor by maximizing the difference between negative and positive volumes. }
\label{fig:notations}
\end{figure*}

\subsection{Preliminaries}
\newcommand{\norm}[1]{\left\lVert#1\right\rVert}
Let a set of tuples $\mathcal{D} = \{(\mathcal{I}, \mathsf{x} )\}$ be the given data consisting of pairs of image $\mathcal{I}$ and its geo-location $\mathsf{x}$. We are interested in learning a mapping function that maps images to feature vectors, ${\phi_\theta:\mathcal{I}\rightarrow \mathsf{f}\in\mathbb{R}^d}$, using mapping parameters $\theta$. In the context of this paper, $\phi$ is a convolutional neural network and $\theta$ are the network parameters. For the task of retrieval-based localization, we wish to learn $\theta$ such that every euclidean distance between features of geometrically close images is minimized. For large scale datasets, measuring the feature distances between all pairs during training may be prohibitively expensive. Therefore, we use the framework of hard example mining, which relies on a set of randomly selected anchor features, say $\mathcal{A}=\{\mathsf{a}\}\subset\mathcal{F}= \{\mathsf{f}\}$. For a given anchor $\mathsf{a}$, the mining process seeks for two sets of so-called positive and negative examples, say $\mathcal{P}$ and $\mathcal{N}$, respectively. Hard example mining further divides positives into easy and hard, say $\mathcal{P}^*=\{\mathsf{p}^*\}$ and $\mathcal{P}^\dag=\{\mathsf{p}^\dag\}$, respectively. Similarly, negatives are composed up of easy negatives $\mathcal{N}^*=\{\mathsf{n}^*\}$ and hard ones $ \mathcal{N}^\dag=\{\mathsf{n}^\dag\}$. In the context of this paper, an example is easy if it is easy to mine whereas a hard example is difficult to correctly distinguish as positive or negative. We further define the volume occupied by positive and negative examples around the anchor $\mathcal{V}^+$ and $\mathcal{V}^-$, respectively. We approximate these volumes by using the parallelotope volume measure~\cite{cai2015law,gover2010determinants}.
Please, refer to Figure~\ref{fig:notations} for a visual illustration of the notations in 2D space.   

\subsection{Mining for Invariance}
Invariance during learning is achieved by minimizing the intra-class variance while maximizing the inter-class variance. One way to improve invariance is through effective mining strategies \cite{angelina2018pointnetvlad} . In fact, the simple strategy of hard negative mining in lazy triplet and quadruplet loss accounts for a large part of the improved invariance learning.
A key aspect of mining-based feature embedding learning is to efficiently create sets $\mathcal{P}$ and $\mathcal{N}$, for each anchor $\mathsf{a}$. Mining these examples, for a large scale localization dataset, poses several engineering challenges. For example, a simple task such as storing all pairwise geometric distances between images---which are used to find true positives or negatives---in memory may not be feasible. 
Needless to say, learning from a large amount of data is necessary for invariance.
To facilitate such learning, we first propose new sampling strategies, namely hard positive and pairwise negative mining, as described below. 

\subsubsection{Hard positives:}
Features for image-retrieval based localization are generally trained with variations of triplet loss.
For each anchor $\mathsf{a}$, feature tuples $(\mathsf{a},\mathcal{P},\mathcal{N})$ are chosen with $|\mathcal{P}|\geq 1$ and $|\mathcal{N}|\geq 1$. The positives $\mathcal{P}$ correspond to images that are geometrically close to the corresponding image of the anchor $\mathsf{a}$, while negatives lie further away.
The triplet loss then imposes the constraint such that the Hausdorff distance $d_{H}(\mathsf{a},\mathcal{P})$ must be smaller than the point-to-set distance $d(\mathsf{a},\mathcal{N})$ at least by some margin. Extensions, such as the weakly supervised ranking~\cite{arandjelovic2016netvlad} (referred to as triplet for brevity), its quadruplet counterpart, and the lazy triplet/quadruplet loss~\cite{angelina2018pointnetvlad} are designed for unknown viewing angles, to circumvent the difficulty of finding true positives during training. These losses therefore replace the Hausdorff distance $d_{H}(\mathsf{a},\mathcal{P})$ by point-to-set distance $d(\mathsf{a},\mathcal{P})$. We argue that this surrogate distance, as a part of precaution, weakens the learning ability of CNNs leading to the failure of matching difficult positive pairs.

In this work, we highlight the \emph{benefit of the Hausdorff distance\footnote{Distance computed between anchor and the image features of its positives.} for learning condition invariant features}. This is rather intuitive as soon as an example of any anchor image and its positives, under strong variations in light, weather, dynamics, or season, are considered. This observation, together with our desire to learn invariant features, lead us to active mining of hard positives $\mathcal{P}^\dag$, for the first time. Among many other benefits, 
condition invariant features offer successful localization using only a small set of reference images. 
It is important to note that the experimental setup that we consider, such as Oxford  RobotCar with hundreds of videos along repeated routes under different conditions, naturally leads to a large number of positives. 
In contrast, for the task of object-instance retrieval (e.g. person/face identification), large 
number of positives may not be available. More precisely, positives may be scarce where labels live in a discrete space (as in identification). Fortunately, this is not the case for localization, where the geometric location is a continuous label. As a matter of fact, our work brings forth the importance
as well as the feasibility of hard positive mining, in the context of learning localization features from large datasets. Note that, we mine both hard positives $\mathcal{P}^\dag$ and easy positives $\mathcal{P}^*$ and combine them to $\mathcal{P}=\mathcal{P}^\dag\cup\mathcal{P}^*$ for training. Our positive sampling strategy is summarized in Algorithm~1.

\subsubsection{Pairwise negatives:}
Naive mining-based loss functions quickly produce zero loss for many tuples, which is addressed by mining hard negative examples. Hard negatives are usually mined from a set of periodically updated cached features.
Given the scale of our training dataset, with many sequences passing through identical places, we notice that simple hard negative mining often results in several very similar hard negatives. For example, a location may be visually similar to another location with many images, such that all these images serve as hard negatives. We therefore, ensure that \emph{each new hard negative is also a negative of all previously selected ones}. We mine both hard and easy negatives, $\mathcal{N}^\dag$ and $\mathcal{N}^*$, and combine them to $\mathcal{N}=\mathcal{N}^\dag\cup\mathcal{N}^*$. For a given anchor $\mathsf{a}$ and a set of cached features $\mathcal{C}$, our negative mining strategy is summarized in Algorithm~2. The pairwise negative mining has one more benefit when used with the objective function proposed in this paper, which will be discussed later in Section~\ref{subSec:volumeLoss}.

 \begin{algorithm}
 \caption{[$\mathcal{P}]=\textbf{HardPos}(\mathsf{a},\mathsf{x}_a,\mathcal{F},\mathcal{X},\mathcal{C})$}
 \begin{algorithmic}[1]
 \State Compute $d_i = d(\mathsf{x}_a,\mathsf{x}_i\in\mathcal{X})$.
 \State Find positives $\hat{\mathcal{P}}\subseteq \mathcal{F}: d_i\leq \alpha$.
 \State Randomly sample $\mathcal{P}^*$ from $\hat{\mathcal{P}}$.
 \State Sort $\hat{\mathcal{C}}=\hat{\mathcal{P}}\cap\mathcal{C}$ based on ${d}(\mathsf{a},\hat{\mathcal{C}})$.
 \State Sample $\mathcal{P}^\dag$ from top-N sorted $\hat{\mathcal{C}}$.
 \State Return $\mathcal{P}=\mathcal{P}^*\cup\mathcal{P}^\dag$.
 \end{algorithmic}
\algrule
Hard positive mining for anchor $\mathsf{a}$ at $\mathsf{x}_a$, cached features $\mathcal{C}$, and all samples $\mathcal{F}$ located at $\mathcal{X=}\{\mathsf{x}\}$.
 \end{algorithm}

 \begin{algorithm}
 \caption{[$\mathcal{N}]=\textbf{PairNeg}(\mathsf{a},\mathsf{x}_a,\mathcal{F},\mathcal{X},\mathcal{C})$}\vspace{0.5mm}
\begin{algorithmic}[1]
 \State Find $\tilde{\mathcal{F}} =\mathcal{F}\setminus\hat{\mathcal{P}}$ using $d(\mathsf{x}_a,\mathcal{X})$.
 \State For $k\leq |\mathcal{N}^\dag|$, from $\tilde{\mathcal{F}}$
 \Statex -- get $\tilde{\mathcal{C}} =\mathcal{C}\cap\tilde{\mathcal{F}}$.
 \Statex -- select hardest $\mathsf{n}_k^\dag$ using $d(\mathsf{a},\tilde{\mathcal{C}})$.
 \Statex -- remove $\mathsf{n}_k^\dag$ and neighbours.
 \State For $j\leq |\mathcal{N}^*|$, from $\tilde{\mathcal{F}}$
 \Statex -- randomly sample $\mathsf{n}_j^*$.
 \Statex -- remove $\mathsf{n}_j^*$ and neighbours.
 \State Return $\mathcal{N}=\mathcal{N}^*\cup\mathcal{N}^\dag$.
\end{algorithmic}
\algrule
Pairwise negative mining for anchor $\mathsf{a}$ at $\mathsf{x}_a$, cached features $\mathcal{C}$, and all samples $\mathcal{F}$ located at $\mathcal{X=}\{\mathsf{x}\}$.
\end{algorithm}

\subsection{Feature Volume-based Objective}\label{subSec:volumeLoss}
In this section, we design an objective function that is used to learn the embedding from the tuple $(\mathsf{a},\mathcal{P},\mathcal{N})$, mined for each anchor $\mathsf{a}$. Let $\mathsf{S}= (\mathsf{s}_1,\ldots,\mathsf{s}_p)$ be the linearly independent random vectors in $\mathbb{R}^d$ with $p\leq d$, represented with respect to the anchor point $\mathsf{a}$ . Then, we make use of the following theorem from random geometry (see \cite{mathai1999random,nielsen1999distribution,cai2015law,gover2010determinants}) to design the target objective function.

\begin{theorem}[Parallelotope Volume Measure~\cite{cai2015law,gover2010determinants}]\label{th:p-para}
A convex hull of $p$ points $\mathsf{S}= (\mathsf{s}_1,\ldots,\mathsf{s}_p)$ in $\mathbb{R}^d$ almost surely determines a p-parallelotope and the volume of this random p-parallelotope is given by $\mathcal{V}_s=\text{det}(\mathsf{S}^\intercal\mathsf{S})^{\frac{1}{2}}$, the square root of the determinant of the random matrix $\mathsf{S}^\intercal\mathsf{S}$. 
\end{theorem}

We learn the feature embedding by approximating the volume occupied by positives and negatives, as the replacements of Hausdorff and point-to-set distances. Recall Figure~\ref{fig:notations}, where positive and negative volumes are approximated around the anchor point. Using Theorem~\ref{th:p-para}, these volumes for 
${\mathsf{P}= (\mathsf{p}_1,\ldots,\mathsf{p}_{|\mathcal{P}|})}$ and 
$\mathsf{N}= (\mathsf{n}_1,\ldots,\mathsf{n}_{|\mathcal{N}|})$, can be expressed as, 
\begin{align}
\begin{split}
 &\mathcal{V}^+ = \text{det}((\mathsf{P}-\mathsf{a})^\intercal(\mathsf{P}-\mathsf{a}))^{\frac{1}{2}}\\
 &\mathcal{V}^- = \text{det}((\mathsf{N}-\mathsf{a})^\intercal(\mathsf{N}-\mathsf{a}))^{\frac{1}{2}}.
 \end{split}
\end{align}

We wish to learn the feature embedding such that for each anchor, the positive volume is minimized while maximizing the negative one. More formally, for the anchor set $\mathcal{A}=\{\mathsf{a}_i\}|_{i=1}^n$, we wish to solve a multi-objective optimization problem formulated as, $\text{min}(\mathcal{V}_i^+, -\mathcal{V}_i^-)$ for all $\mathsf{a}_i\in\mathcal{A}$. Optimizing such an objective function is not only difficult, but is also often intractable. Therefore, we reformulate a surrogate optimization problem, while also omitting the square roots, as,

\begin{align}\label{eq:volumeCost}
\begin{split}
&\min \sum_{\mathsf{a}_i\in\mathcal{A}}{\bigg((\mathcal{V}_i^+)^2-(\mathcal{V}_i^-)^2 \bigg)}, \\
&\text{s.t.} \,\,\norm{\mathsf{f}_j}=1, \forall \mathsf{f}_j\in\cup_{i=1}^n(\mathcal{A}_i\cup\mathcal{N}_i\cup\mathcal{P}_i) .
\end{split}
\end{align}

 We realize a neural network $\phi_\theta:\mathcal{I} \rightarrow \mathsf{f}\in\mathbb{R}^s$
 that maps images into feature space. The network is trained for parameters $\theta$ such that the feature embedding is optimized for the problem of~\eqref{eq:volumeCost}. The constraint of~\eqref{eq:volumeCost} is enforced simply by normalizing all features to unit norm. The square volumes $(\mathcal{V}_i^+)^2$ and $(\mathcal{V}_i^-)^2$ are computed as the products of eigenvalues of the matrices $(\mathsf{P}-\mathsf{a})^\intercal(\mathsf{P}-\mathsf{a})$ and $(\mathsf{N}-\mathsf{a})^\intercal(\mathsf{N}-\mathsf{a})$, respectively. In our implementation, we compute the volumes after projecting the features down to $r$ dimensions, with $r$ smaller than $s$, $|\mathcal{P}|$, and $|\mathcal{N}|$. In this regard, our pairwise negative mining strategy, presented above, turns out to be very useful to ensure non-zero negative volumes in high dimensions. Please, find more details regarding our implementation in Section~\ref{sec:implementation_details}. 
 
In the following, we provide some theoretical insights of our objective function before proceeding to the implementation details and experimental evaluations.
\begin{remark}
The maximum squared-volume of a p-paral\-lelotope in a lower dimension, under a linear projection model, can be computed using the product of the eigenvalues of the matrix $\mathsf{S}^\intercal\mathsf{S}$. This is done by considering only the most significant eigenvalues of the same number as that of the dimension to be projected to.
\end{remark}
\begin{remark}
Minimizing the volume $\mathcal{V}^+$ has the general tendency to minimize the Hausdorff distance $d_H(\mathsf{a},\mathcal{P})$ between anchor and positive examples.
\end{remark}
\begin{remark}
Maximizing the volume $\mathcal{V}^-$ has the general tendency to maximize the point-to-set distance $d(\mathsf{a},\mathcal{N})$ between anchor and negative examples.
\end{remark}
There are four major benefits of using feature volume-based objective functions:  (i) features make use of all the dimensions of the embedding, thus avoiding collapsing into low-dimensional space; (ii)  suitable for large $\mathcal{P}$ or $\mathcal{N}$, as in our case of learning from large scale data, because the eigenvalues of $\mathsf{S}^\intercal\mathsf{S}$  can still be computed efficiently using $\mathsf{S}\mathsf{S}^\intercal$, if $r\ll |\mathcal{P}|$ or $r\ll |\mathcal{N}|$; (iii) offers quicker convergence during learning; (iv) provides better test accuracy.

\section{Implementation Details}\label{sec:implementation_details}
\paragraph{Datasets.}
We conduct experiments on three publicly available real world datasets--COLD Freiburg~\cite{pronobis2009cold}, Oxford RobotCar~\cite{maddern20171}, and Pittsburgh (Pitts250k)~\cite{torii2013visual}.
The COLD Freiburg dataset consists of two geographically disjoint parts, A and B. We use part A for training and part B for testing. For Oxford RobotCar, we manually divide the map into a test region and two training regions with different sizes, as shown in Figure \ref{fig:test_sets}. Again, we ensure that training and testing regions do not overlap or share covisible buildings.
In its original form, \textit{Pittsburgh} contains 250k reference images. For efficient testing, we ignore all reference images that are more than 100 metres away from the nearest query and remove all images with pitch 2 (i.e.\ images that mostly show sky). The locations of the remaining images are shown in Figure~\ref{fig:test_sets}. The evaluations with features trained on Pittsburgh are obtained using the publicly available weights of \cite{arandjelovic2016netvlad}, which were trained on a 30k subset of Pittsburgh. Finally, we also report results based on features that are not trained on any localization dataset. Instead, the weights are initialized with features trained for ImageNet~\cite{deng2009imagenet} classification. Table~\ref{tab:training_size} summarizes the number of training images for the five different training setups. For clarity, we use Roman numerals to identify the different training sets. For reproducibility, we will include a list of all training and test images on our project website. 

\paragraph{Choice of Reference Images.}
During localization, query images are matched against a database of reference images. 
The choice of reference images is based on two assumptions. Firstly, the reference storage is limited, which is why we only select a limited number of reference images. Secondly, it is justified to choose the easiest condition as a reference, given that anyone designing an image-retrieval based localization system will most likely have control over the initial reference conditions. In that light, we choose a cloudy reference for Freiburg and an overcast reference sequence for Oxford RobotCar. These conditions have less glare than sunny sequences and better light than night images. 
Figure~\ref{fig:test_sets} shows the selected reference locations.

\paragraph{Data Prepossessing.}
\label{sec:data_preprocessing}
For training on Cold Freiburg (I) we do not preprocess the data. For training on the smaller Oxford RobotCar (II) region, the training sequences were selected based on the visual assessment of their INS trajectories. When training on the larger Oxford RobotCar (III) region, data preprocessing becomes more important, especially when learning invariances by training with hard positives.
Given that the Oxford RobotCar dataset is somewhat noisy, we filter out bad or atypical locations by removing location outliers and exclude under and over exposed images. For the sake of reproducibility, we provide a list of all included images on our project website.

\paragraph{Network Architecture.}
We use a VGG-16~\cite{Simonyan2014} network cropped at the last convolutional layer and extend it with a NetVLAD~\cite{arandjelovic2016netvlad} layer as implemented by~\cite{cieslewski2018data}, initializing the network with off-the-shelf ImageNet~\cite{deng2009imagenet} classification weights, i.e.\ weights that have not yet been retrained for localization. We also test a very simple alternative to NetVLAD spatial pooling, which simply flattens the output of the last convolutional layer of VGG-16 into a feature vector. This type of feature requires that input images are consistently of the same size. We achieve this by scaling and cropping. Experiments using this type of simplified features are marked with an asterisk.

\paragraph{Training.} All models are trained on a single Nvidia TitanX GPU. 
For training with Cold Freiburg (I) and the smaller Oxford RobotCar set (II), we use the implementation from~\cite{angelina2018pointnetvlad}, adapted to handle images instead of point clouds, and the training parameters specified in~\cite{arandjelovic2016netvlad}, reducing the learning rate to 0.000001, the number of queries per batch to two, and the number of positives and negatives per query to six each.
With this smaller query size it becomes possible to train VGG-16 in its entirety and not only down to conv5 layer as it is done by~\cite{arandjelovic2016netvlad}.
During training, for each epoch, we iterate over all training images.
For each image, we sample positives from within a radius of $r_1$, and negatives that are at least $r_2$ distance away.
We set $r_1$ to 1 meter for COLD Freiburg and, in accordance with~\cite{arandjelovic2016netvlad}, to 10 meters for Oxford RobotCar. 
$r_2$ is set to 4 meters and 25 meters respectively. 
Additionally, for COLD Freiburg, we exclude images with a difference in yaw angle larger than 30 degrees.
For Oxford RobotCar this is not necessary, as the region we work with, only contains few images which are geometrically close but not co-visible.
For each image, half of the negatives are chosen via hard negative mining. 
We update our feature cache for hard negative mining every 400 iterations for COLD Freiburg and every 1000 iterations for Oxford RobotCar.

For efficient training on the larger Oxford RobotCar training set (III) we re-implement the training procedure. 
Hard negative and positive mining as well as tuple assembly are now handled by multiple threads in parallel.
Unlike Oxford II, Oxford III does contain intersecting sequences.
We therefore exclude images with a difference in yaw angle larger than 30 degrees.
This is important when training with hard positives.
Given that Oxford RobotCar (III) contains numerous sequences which run through the same locations, but not every location is visited the same number of times, we redefine the concept of one epoch as having gone over each location along the standard driving route.

\paragraph{Losses.}
We compare our volume loss introduced in Section \ref{subSec:volumeLoss} to the triplet loss version of~\cite{arandjelovic2016netvlad}, quadruplet loss~\cite{chen2017beyond}, lazy triplet and quadruplet loss~\cite{angelina2018pointnetvlad}, and triplet loss with (Huber) distance~\cite{thoma2020geometrically}. We also modify the triplet and quadruplet loss to better benefit from hard positive mining. Instead of using the minimum distance between anchor and all its positives we use the Hausdorff distance. Please note that our volume loss also uses and benefits from hard positive and pairwise negative mining as empirically demonstrated in Section \ref{sec:experiments}.

\paragraph{Evaluation metric.}
We consider an image to be correctly localized if the distance between the top-1 retrieved reference and query location is smaller than a given distance threshold $d$. If not stated otherwise, this threshold is set to 10m for outdoor evaluations and 1m for Cold Freiburg, which are also the maximum positive radii used during training. For any given testing condition, we report the percentage of correctly localized images, i.e.\ the localization accuracy. 

\begin{figure*}[t]
     \centering
     \includegraphics[width=\textwidth]{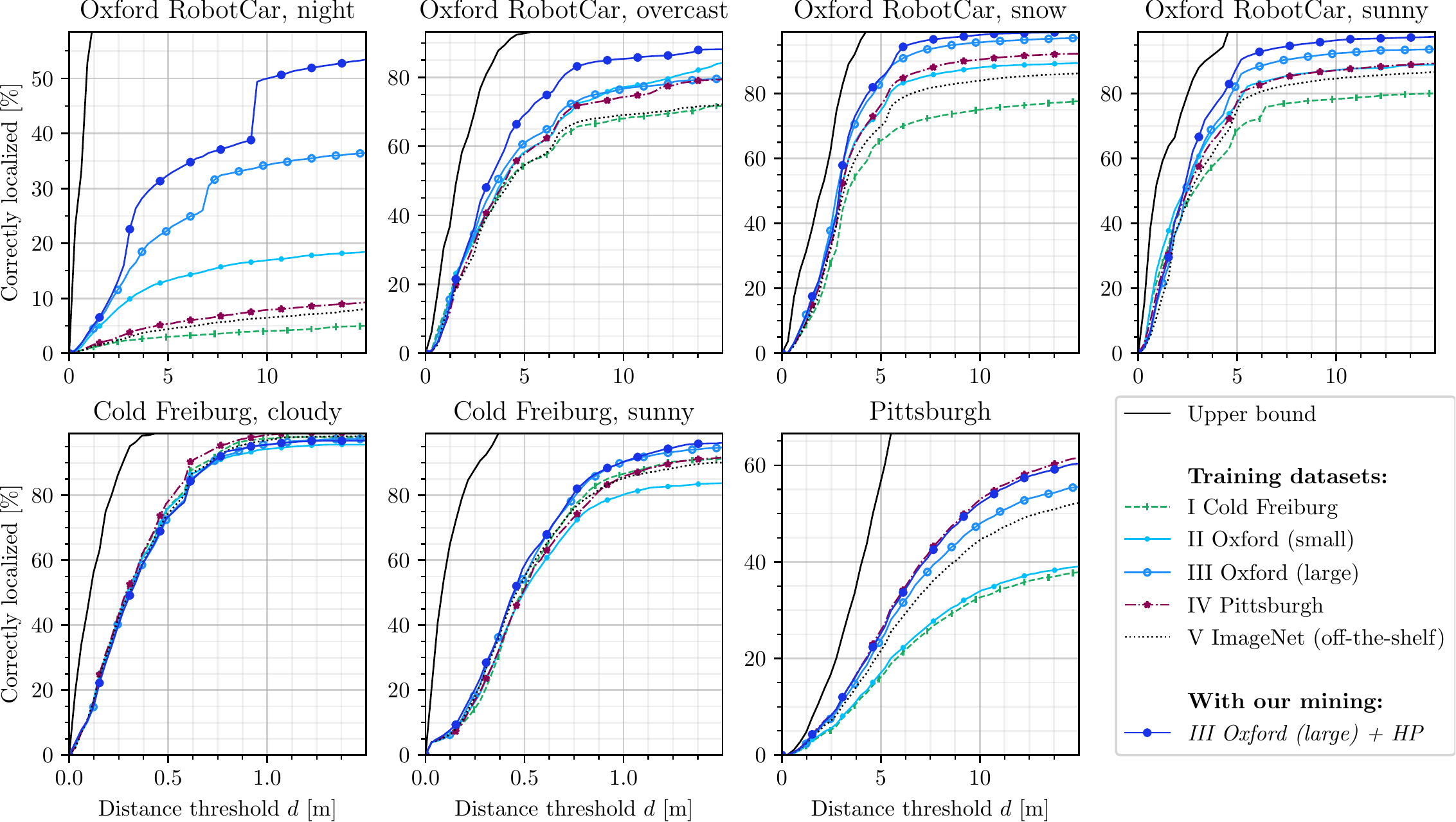}
     \caption{Localization accuracy as a function of distance threshold $d$ for models trained with triplet loss on training regions I-V. To evaluate feature generalization, all models are evaluated on 7 different conditions from Oxford RobotCar, Cold Freiburg and Pittsburgh. The upper bound is given by the minimal distance of all queries to their geometrically closest reference. 
     The blue curve with solid circle markers shows the condition invariance boost obtained with our hard positive mining (HP).}
     \label{fig:roc_datasets}
 \end{figure*}

\begin{table*}
\begin{center}
\resizebox{\textwidth}{!}{%
\begin{tabular}{@{}clcccccccc@{}}
\toprule
\multicolumn{2}{l}{}  & \multicolumn{4}{c}{Oxford RobotCar} & \multicolumn{2}{c}{Cold Freiburg} & Pittsburgh & Mean  \\
 &
  Loss &
  Night &
  Overcast &
  Snow &
  Sunny &
  Cloudy &
  Sunny &
   &
   \\\midrule &
  Thresholds [m] &
  5.0/10.0/15.0 &
  5.0/10.0/15.0 &
  5.0/10.0/15.0 &
  5.0/10.0/15.0 &
  0.5/1.0/1.5 &
  0.5/1.0/1.5 &
  5/10/15 & \\
  &
Upper bound &
  100.0/100.0/100.0 &
  92.7/99.3/100.0 &
  99.9/100.0/100.0 &
  100.0/100.0/100.0 &
  100.0/100.0/100.0 &
  100.0/100.0/100.0 &
  57.3/96.8/98.0 &
  92.8/99.4/99.7  \\\midrule
  
\multirow{6}{*}{I} &
Triplet \cite{schroff2015facenet} &
  3.0/4.0/5.0 &
  54.7/68.1/71.7 &
  65.5/75.0/77.6 &
  68.7/78.3/80.1 &
  52.1/86.7/91.3 &
  76.8/97.5/98.1 &
  16.2/32.7/37.9 &
  48.1/63.2/66.0 \\
  &
Quadruplet \cite{chen2017beyond} &
  3.4/4.6/6.1 &
  55.2/69.3/73.7 &
  70.3/81.5/83.8 &
  74.3/84.7/86.0 &
  54.3/84.8/88.7 &
  76.3/97.0/98.0 &
  18.3/36.2/42.5 &
  50.3/65.5/68.4 \\
  &
Lazy triplet \cite{angelina2018pointnetvlad} &
  3.1/4.5/5.2 &
  41.1/57.8/61.3 &
  57.5/69.3/71.6 &
  64.9/75.9/77.3 &
  50.8/82.4/86.1 &
  77.2/96.6/97.7 &
  18.0/35.4/41.0 &
  44.7/60.3/62.9 \\
  &
Lazy quadruplet \cite{angelina2018pointnetvlad} &
  3.6/5.3/6.3 &
  56.5/70.9/76.5 &
  64.9/75.4/77.6 &
  70.6/79.4/81.1 &
  52.3/81.4/86.1 &
  75.7/95.3/96.4 &
  16.9/34.2/39.6 &
  48.6/63.1/66.2 \\
  &
Triplet + distance \cite{thoma2020geometrically} &
  5.2/6.7/7.8 &
  48.5/66.9/74.3 &
  63.3/75.2/78.3 &
  70.5/78.4/80.2 &
  52.7/78.7/82.0 &
  \textbf{79.6}/97.9/98.5 &
  19.3/37.8/44.1 &
  48.4/63.1/66.5 \\
  &
Triplet + Huber dist.~\cite{thoma2020geometrically} &
  4.0/6.4/8.4 &
  51.5/69.3/73.4 &
  67.6/78.0/80.0 &
  68.8/78.8/80.6 &
  53.2/81.2/84.5 &
  78.5/97.0/97.3 &
  18.6/37.3/43.6 &
  48.9/64.0/66.8 \\\midrule
\multirow{5}{*}{II} &
  Triplet \cite{schroff2015facenet} &
  13.3/17.0/18.5 &
  58.3/77.1/84.2 &
  75.5/88.0/89.4 &
  77.2/87.3/89.0 &
  50.3/80.3/83.8 &
  76.2/94.3/95.6 &
  17.1/33.9/39.1 &
  52.6/68.3/71.3 \\
  &
Quadruplet \cite{chen2017beyond} &
  22.0/26.1/27.3 &
  54.5/72.7/77.6 &
  71.1/80.3/81.9 &
  75.1/87.2/89.0 &
  48.5/77.9/81.4 &
  74.8/94.7/96.0 &
  18.3/36.6/41.9 &
  52.0/67.9/70.7 \\
  &
Lazy triplet \cite{angelina2018pointnetvlad} &
  14.2/17.4/18.3 &
  61.4/76.1/78.3 &
  74.7/83.0/84.6 &
  69.3/76.4/78.1 &
  52.7/81.6/85.7 &
  74.9/96.4/97.6 &
  17.0/33.0/38.0 &
  52.0/66.3/68.7 \\
  &
Lazy quadruplet \cite{angelina2018pointnetvlad} &
  15.1/19.1/20.2 &
  57.8/72.7/76.4 &
  76.0/87.5/89.0 &
  72.1/81.6/82.7 &
  45.8/74.5/78.7 &
  73.4/91.8/92.8 &
  18.0/35.7/41.3 &
  51.2/66.1/68.7 \\
  &
Triplet + Huber dist.~\cite{thoma2020geometrically} &
  10.8/13.7/15.6 &
  59.7/74.1/76.5 &
  75.8/88.0/90.1 &
  77.7/84.9/86.0 &
  48.3/79.1/83.8 &
  74.0/93.2/94.0 &
  15.6/29.8/34.7 &
  51.7/66.1/68.7 \\\midrule
\multirow{10}{*}{III} &
 Triplet \cite{schroff2015facenet} &
  22.7/34.3/36.4 &
  60.9/76.8/79.6 &
  83.5/95.8/97.2 &
  84.6/92.4/93.7 &
  55.2/90.3/94.8 &
  73.2/95.7/97.4 &
  24.0/47.9/55.5 &
  57.7/76.2/79.2 \\
  &
Quadruplet \cite{chen2017beyond} &
  20.2/30.5/33.8 &
  59.2/76.0/78.6 &
  82.1/94.9/96.5 &
  84.8/92.2/94.1 &
  57.7/86.9/91.7 &
  73.8/96.0/96.7 &
  23.4/46.6/53.8 &
  57.3/74.7/77.9 \\
  &
Lazy triplet \cite{angelina2018pointnetvlad} &
  26.9/33.1/35.8 &
  65.0/80.7/83.0 &
  83.0/95.5/96.5 &
  82.8/93.5/95.3 &
  57.2/88.1/93.6 &
  76.8/96.9/97.8 &
  24.0/48.5/56.3 &
  59.4/76.6/79.7 \\
  &
Lazy quadruplet \cite{angelina2018pointnetvlad} &
  28.1/41.7/44.6 &
  58.7/74.2/76.5 &
  83.8/97.0/97.8 &
  83.3/89.8/91.3 &
  55.6/88.6/92.9 &
  74.7/96.3/98.5 &
  25.2/50.6/58.2 &
  58.5/76.9/80.0 \\
  &
Triplet + distance \cite{thoma2020geometrically} &
  27.1/32.4/34.0 &
  66.7/82.2/84.6 &
  82.3/96.2/97.4 &
  82.9/93.4/94.7 &
  57.1/86.3/91.8 &
  73.5/95.3/96.5 &
  20.6/40.6/46.8 &
  58.6/75.2/78.0 \\
  &
Triplet + Huber dist.~\cite{thoma2020geometrically} &
  31.6/42.9/45.3 &
  56.1/74.0/76.5 &
  83.8/95.9/97.2 &
  84.2/92.6/93.9 &
  56.7/87.3/92.5 &
  74.3/96.1/96.9 &
  21.6/43.6/50.4 &
  58.3/76.1/78.9 \\
  &
\textit{Triplet + HP} &
  32.4/50.0/53.5 &
  \textbf{69.3}/\textbf{85.4}/88.1 &
  84.8/98.1/99.1 &
  89.6/96.4/97.5 &
  \textbf{58.7}/\textbf{90.4}/\textbf{96.2} &
  73.5/95.6/96.9 &
  25.4/52.2/60.4 &
  61.9/\textbf{81.2}/\textbf{84.5} \\
  &
\textit{Quadruplet + HP} &
  25.9/37.9/47.3 &
  66.4/84.2/\textbf{89.1} &
  85.0/97.1/98.1 &
  86.1/96.0/97.5 &
  54.2/90.1/93.4 &
  76.2/97.2/99.1 &
  26.3/54.0/62.5 &
  60.0/79.5/83.9 \\
  &
\textit{Volume without HP} &
  32.6/38.7/40.9 &
  63.2/79.0/83.5 &
  87.3/98.5/99.0 &
  83.1/95.7/96.7 &
  53.8/87.1/93.1 &
  73.2/97.0/98.4 &
  27.7/56.6/65.7 &
  60.1/78.9/82.5 \\
  &
$\mathit{Volume}^*$ without HP &
  \textbf{52.9}/\textbf{62.6}/\textbf{65.5} &
  64.2/76.9/80.8 &
  87.2/97.5/98.8 &
  88.8/\textbf{97.4}/98.0 &
  49.8/83.1/88.0 &
  66.6/91.6/92.9 &
  26.2/53.0/61.0 &
  \textbf{62.3}/80.3/83.6 \\
  &
\textit{Volume} &
  35.0/43.6/46.2 &
  66.3/84.6/88.7 &
  87.5/\textbf{98.7}/\textbf{99.2} &
  87.2/96.0/97.0 &
  54.6/88.9/93.3 &
  76.5/\textbf{99.0}/\textbf{99.5} &
  \textbf{28.2}/\textbf{57.3}/\textbf{66.1} &
  62.2/81.1/84.3 \\
  &
$\mathit{Volume}^*$ &
  48.9/57.9/60.3 &
  65.9/79.6/82.9 &
  \textbf{87.7}/97.9/98.9 &
  \textbf{89.9}/97.3/\textbf{98.1} &
  49.3/83.6/88.7 &
  67.5/91.2/93.2 &
  26.2/53.2/61.1 &
  62.2/80.1/83.3 \\\midrule
IV &
 Triplet &
  5.4/7.9/9.3 &
  57.8/74.3/79.4 &
  76.9/90.5/92.3 &
  79.6/87.2/89.3 &
  51.4/85.8/91.7 &
  78.1/98.6/99.4 &
  26.0/52.9/61.5 &
  53.6/71.0/74.7 \\\midrule
V &
Off-the-shelf \cite{deng2009imagenet} &
  4.4/6.5/8.0 &
  54.5/69.1/72.1 &
  70.2/84.0/86.3 &
  76.8/84.7/86.6 &
  55.2/85.3/90.1 &
  74.4/97.0/98.2 &
  21.7/44.6/52.2 &
  51.0/67.3/70.5\\\bottomrule
\end{tabular}%
}
\caption{Localization accuracy SOTA comparison for 23 different training setups and 7 query sequences. Numerals I-V indicate the training region (I: Cold Freiburg, II: Oxford RobotCar (small), III: Oxford RobotCar (large), IV: Pittsburgh30K, V: ImageNet). Training and testing regions are always geographically disjoint. The table also reports the mean performance of each training setting over all seven test conditions. The best value for each testing condition and threshold is marked in bold.}
\label{tab:quantitative}
\end{center}
\end{table*}

\begin{figure*}
\begin{tabularx}{\textwidth}{C{1} C{1} C{1} C{1} C{1} C{1} C{1} C{1} }\toprule
\multicolumn{4}{c}{Volume}& \multicolumn{4}{c}{Triplet}\\\midrule
\multicolumn{2}{c}{Image}& \multicolumn{2}{c}{Grad-CAM}&
\multicolumn{2}{c}{Image}& \multicolumn{2}{c}{Grad-CAM}\\
Query & Retrieved & Query & Retrieved & Query & Retrieved & Query & Retrieved\\\midrule
\multicolumn{4}{c}{\includegraphics[width=0.45\textwidth]{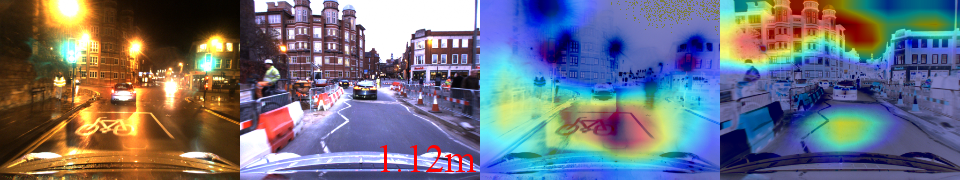}}&
\multicolumn{4}{c}{\includegraphics[width=0.45\textwidth]{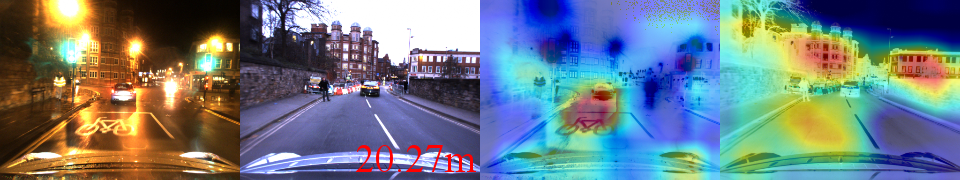}}\\
\multicolumn{4}{c}{\includegraphics[width=0.45\textwidth]{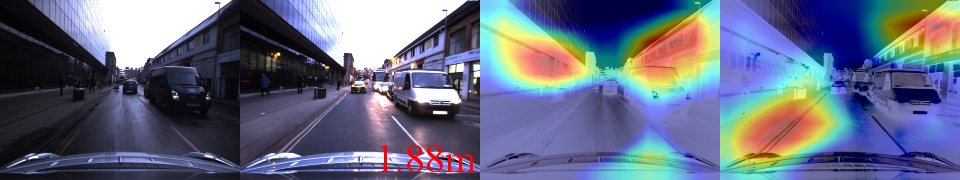}}&
\multicolumn{4}{c}{\includegraphics[width=0.45\textwidth]{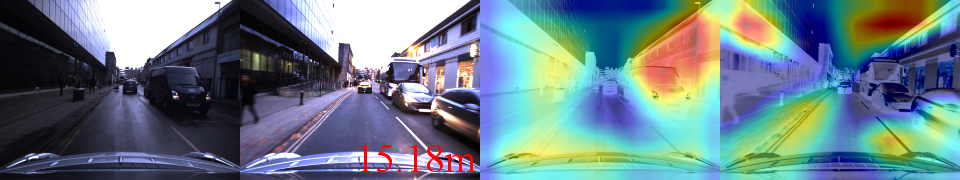}}\\
\multicolumn{4}{c}{\includegraphics[width=0.45\textwidth]{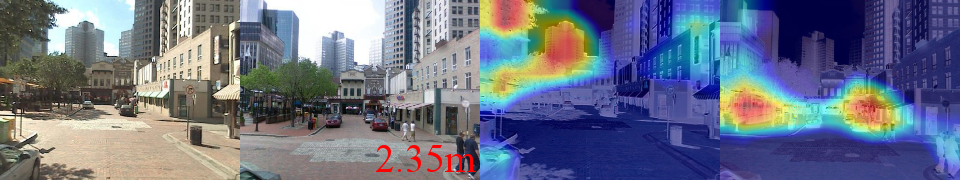}}&
\multicolumn{4}{c}{\includegraphics[width=0.45\textwidth]{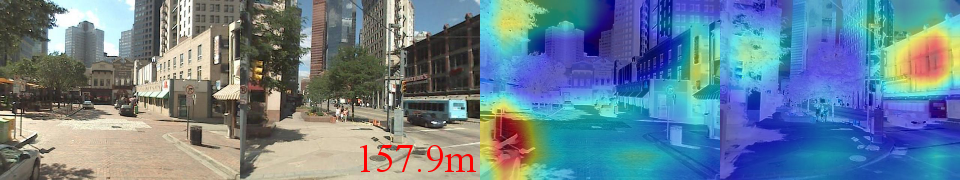}}\\\bottomrule
\end{tabularx}
\caption{Selected visual examples comparing our volume loss to triplet loss (both trained on the large Oxford RobotCar (III) training region). For each query, we show the top-1 retrieved reference with localization error and Grad-CAM~\cite{selvaraju2017grad} representation, highlighting the image regions which contributed to the retrieval. The queries are taken from Oxford night, snow, and Pittsburgh. More visual results in supplementary video.}
\label{tab:visuals}
\end{figure*}

 \begin{figure*}[p]
     \centering
     \includegraphics[width=\textwidth,trim={0 1.2cm 0 0},clip]{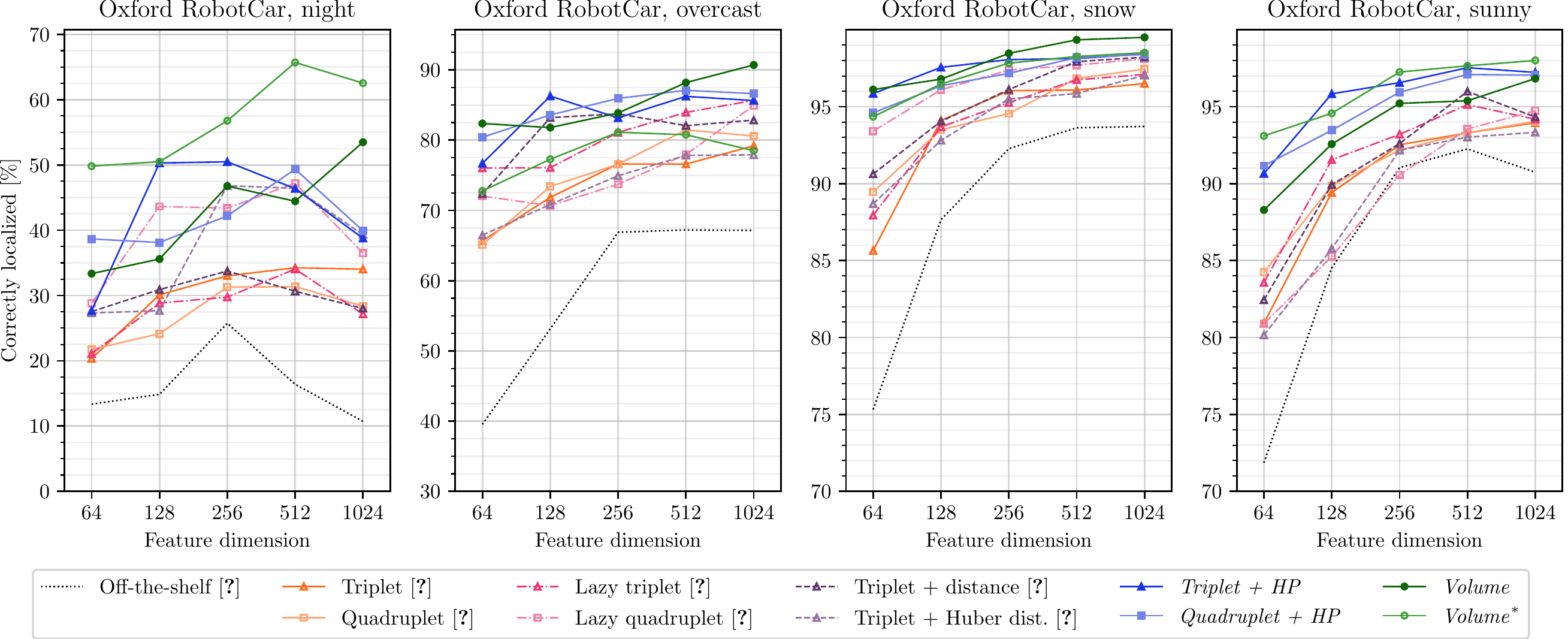}
     \includegraphics[width=\textwidth,trim={0 1.2cm 0 0},clip]{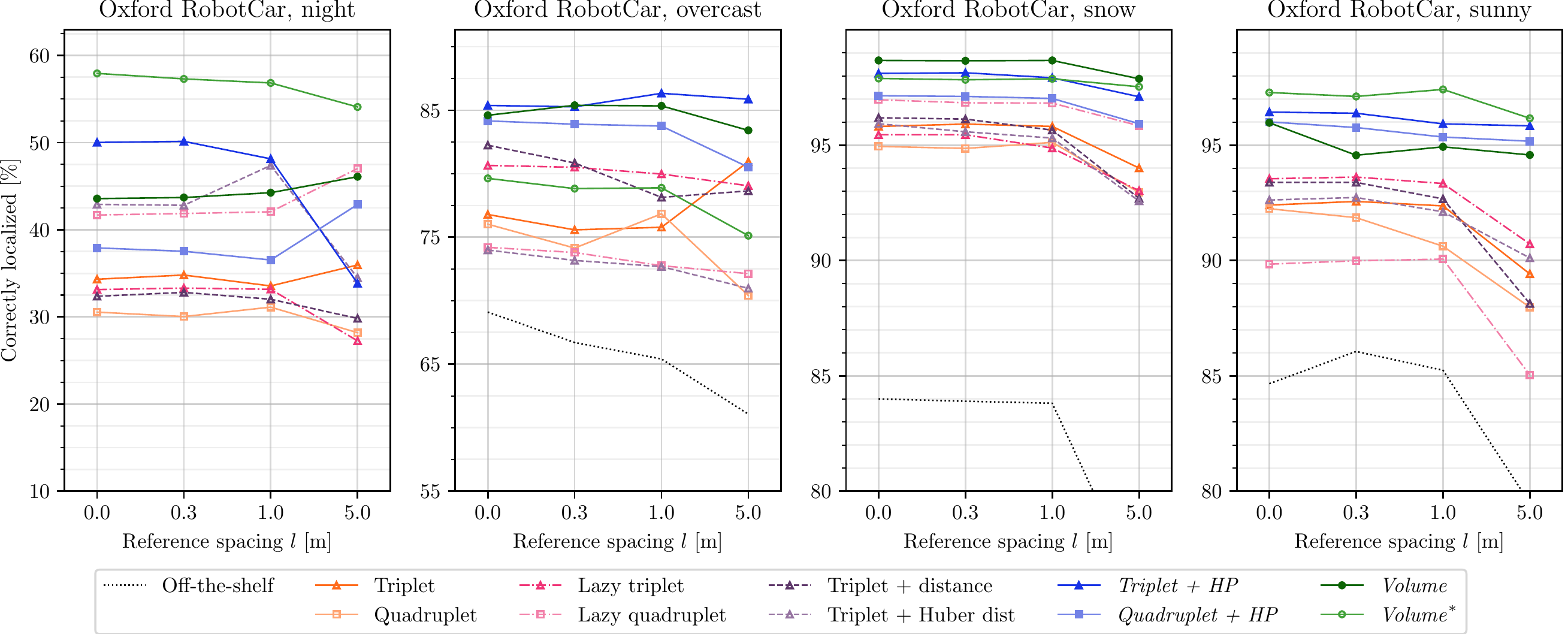}
    \includegraphics[width=\textwidth,trim={0 0 0 0},clip]{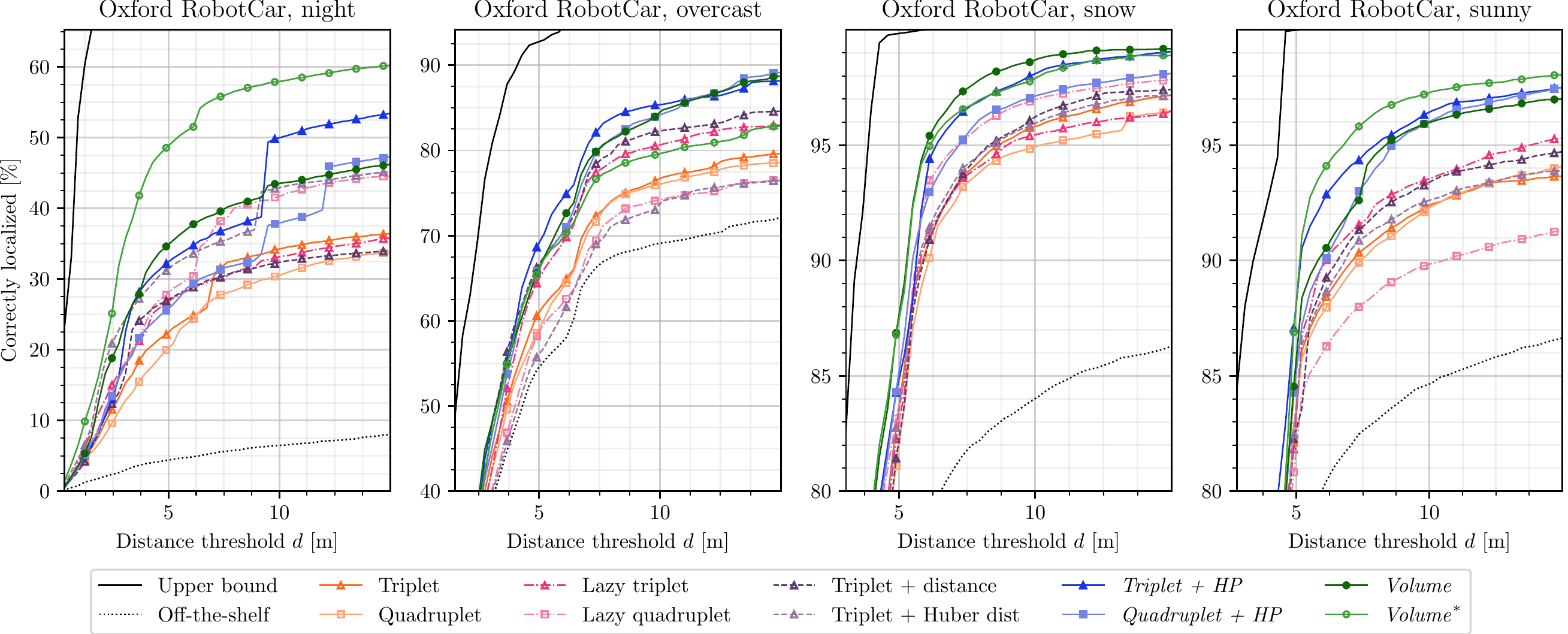}
     \caption{Localization accuracy SOTA comparison for models trained on large Oxford RobotCar (III) as a function of {\textit{1) Feature dimension (top):}}  The number of dimensions after PCA with whitening. {\textit{2) Reference spacing (middle):}} The distance $l$ between two consecutive reference images. For $l=0.0$ all images of the reference sequence are used. The resulting reference set sizes are 6218, 3707, 1405, and 335. {\textit{3) Distance threshold (bottom):}} Maximum distance $d$ between ground truth query location and retrieved reference location for which images are considered correctly localized. If not stated otherwise, $\text{dim}=256$, $l=0.0$ and $d=10.0$.
     All models use VGG-16 with NetVLAD~\cite{arandjelovic2016netvlad} spatial pooling, except volume*, which directly works on the last convolutional output of VGG-16. {\textit{Triplet + HP}} and {\textit{quadruplet + HP}} highlight the benefits of our mining for invariance, while {\textit{volume}} and {\textit{volume*}} are obtained with our novel volume-based objective. For more results, see supplementary video and appendix \ref{app:app}.
     }
     \label{fig:roc}
 \end{figure*}

\section{Experiments}
\label{sec:experiments}
In Table~\ref{tab:quantitative} we report results on 23 different training settings, 21 of which we have trained ourselves on the Cold Freiburg and Oxford RobotCar datasets. Each training setting is evaluated on 7 different test sequences, shown in Fig.~\ref{fig:test_sets}. The numbers are reported for three different distance thresholds $d$, 5m/10m/15m for outdoor datasets and 0.5m/1m/1.5m for Cold Freiburg. Overall, our volume loss trained on 1M+ Oxford RobotCar images (III) outperforms all other methods. Please note that the volume loss also uses hard positive and pairwise negative mining. 
Figure~\ref{fig:roc_datasets} shows that models trained on the largest and most diverse data set, Oxford RobotCar (III), tend to also generalize to other datasets, while models trained on a dataset without night conditions, predictably fail on the challenging Oxford RobotCar night evaluation sequence. It can also be observed that hard positive mining improves localization and generalization. 
Figure~\ref{fig:roc} shows the performance of compared SOTA methods and our models trained on Oxford RobotCar (III) for varying feature dimensions, reference set sizes and localization threshold $d$. The middle row shows that the performance of volume loss trained models degrades very gracefully when the number of reference images are reduced drastically, from over 6000 to 335. An obvious advantage of a sparse reference set is a smaller memory footprint and faster retrieval. 
Finally, we report visual results in Figure~\ref{tab:visuals}. More results and ablation studies can be found in the supplementary video and appendix~\ref{app:app}.

\section{Discussion}
\label{sec:discussion}
\paragraph{Size matters.} Training set size is a crucial factor when learning for localization in new regions. If the goal is to localize on the training region itself, a small set suffices. But generalizability to new regions or even datasets, requires a large and diverse training dataset, as our experiments clearly reveal. 

\paragraph{Diversity matters.} The training set should be as diverse as possible. 
Query conditions that are not covered in the training set will most likely result in poor retrieval performance for that specific condition during deployment.

\paragraph{Hard positive mining helps.}
Hard positive mining is essential to efficiently learn to localize on difficult conditions, such as night images. 

\paragraph{Fine-tuning helps.}
If at all possible, it is always beneficial to fine-tune directly on the region where one is planning to localize. Alternatively, training on images with similar appearance (i.e.\ same dataset but a geographically disjoint region or a different dataset but similar conditions) also helps.

\paragraph{Are we better than ImageNet Pre-training?} Yes. For the task of localization. Our features learned on the large Oxford RobotCar (III) training set with volume loss and hard positive mining outperform the off-the-shelf ImageNet pre-trained features on all seven test regions. It is fair to assume, this will also be the case for further localization applications. We will therefore provide our trained weights and models on our project website. 

\section{Conclusion}
In this paper, we compared the performance of several deep features for the task of image retrieval based localization. More importantly, the comparisons were carried out after training each of those features on a dataset of more than a million images.
We have analyzed the various cases of illumination, weather, and content, across three diverse datasets to provide valuable insights with extensive and large scale accuracy measure quantification.
Our experiments clearly show that training on a large scale dataset is necessary if the features are desired to work on difficult conditions and to generalize to new regions and datasets. 
We have also introduced (i) a feature volume-based loss function and (ii) hard positive as well as pairwise negative mining strategies, which greatly boost the performance on difficult conditions and make the learned features generalizable.
The source code and the learned models presented in this paper are publicly available\footnote{\url{https://github.com/janinethoma/learning1M}}.

\bibliographystyle{ieee_fullname}
\bibliography{biblio}

\renewcommand\thefigure{\Roman{figure}}   
\setcounter{figure}{0}   
\appendix
\section{Appendix}
\label{app:app}

This appendix contains ablation studies \ref{app:ablation}, additional quantitative results \ref{app:quantitative} and feature visualizations \ref{app:feature_visualizations}. For a comprehensive collection of visual results, please see our supplementary video at on our project website.

\subsection{Ablation Studies}
\label{app:ablation}
\subsubsection{Spatial Pooling}
\begin{figure*}
    \centering
    \includegraphics[width=0.9\textwidth]{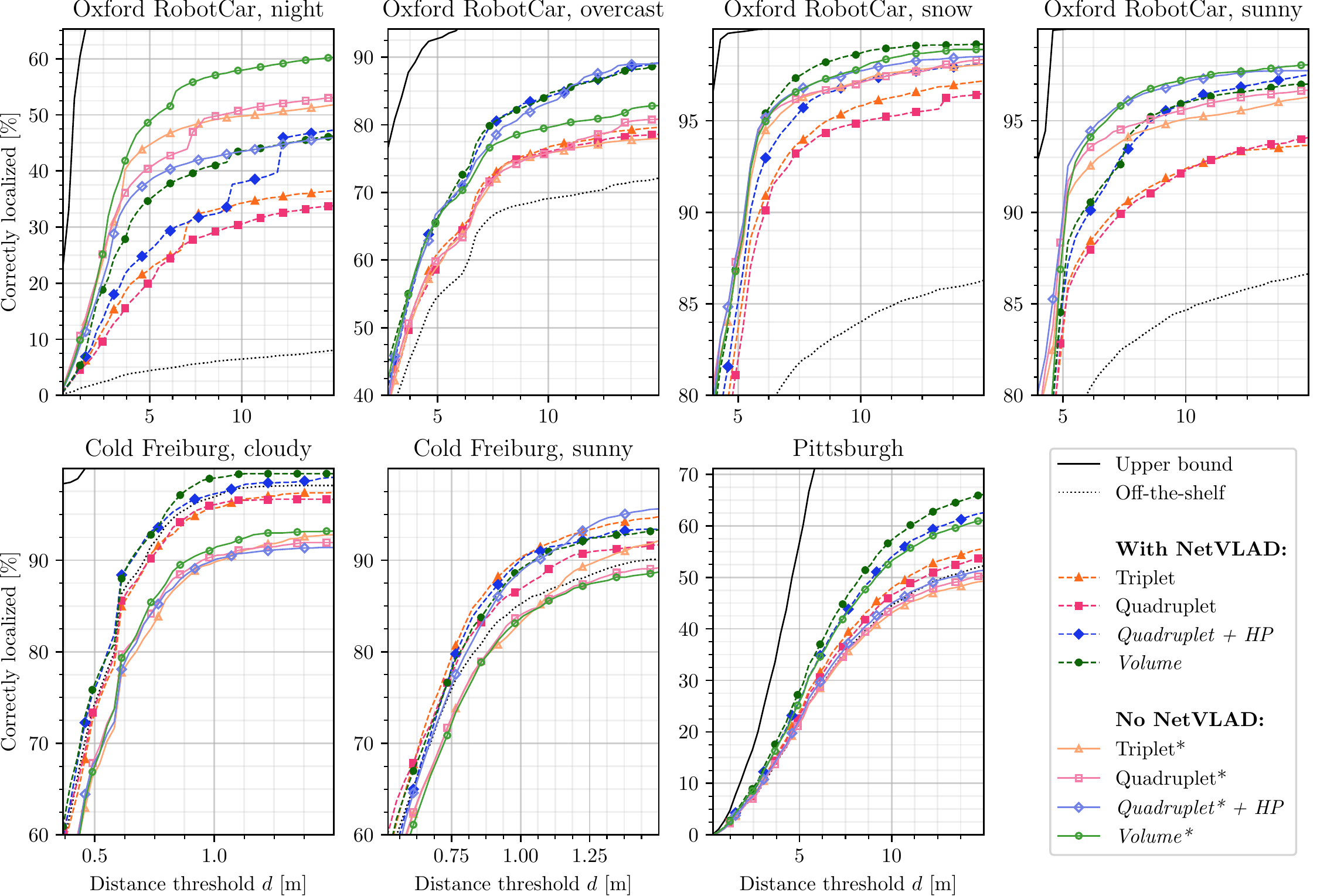}
    \caption{Localization accuracy comparison with and without NetVLAD spatial pooling for models trained on large Oxford RobotCar (III).}
    \label{fig:pooling}
\end{figure*}

As described in Section~\ref{sec:implementation_details}, we assess the necessity of Net\-VLAD \cite{arandjelovic2016netvlad} by comparing it to a simple baseline. This baseline, which we denote with an asterisk, flattens the output of the last VGG-16 convolutional layer into a feature vector. Figure~\ref{fig:pooling} shows the resulting localization accuracy for four different losses: triplet~\cite{arandjelovic2016netvlad}, quadruplet~\cite{chen2017beyond}, quadruplet with our mining strategies, and our volume loss.
It can be observed that, for the Oxford RobotCar night and sunny conditions, dropping NetVLAD leads to a considerable performance boost. We believe that this boost is due to the preservation of all global spatial relations in the features calculated without NetVLAD. However, Figure~\ref{fig:pooling} also shows that our naive baseline does not generalize well to sequences captured with different image sizes and cameras (Cold Freiburg and Pittsburgh). Please note that for both pooling strategies, our hard-positive mining and our volume-based loss boost the localization performance. It is therefore justified to assume, that the benefits of our contributions will also apply to different pooling strategies such as R-MAC~\cite{tolias2015particular} and its variants~\cite{Gordo2017,magliani2018accurate}.

\begin{figure*}
    \centering
    \includegraphics[width=0.9\textwidth]{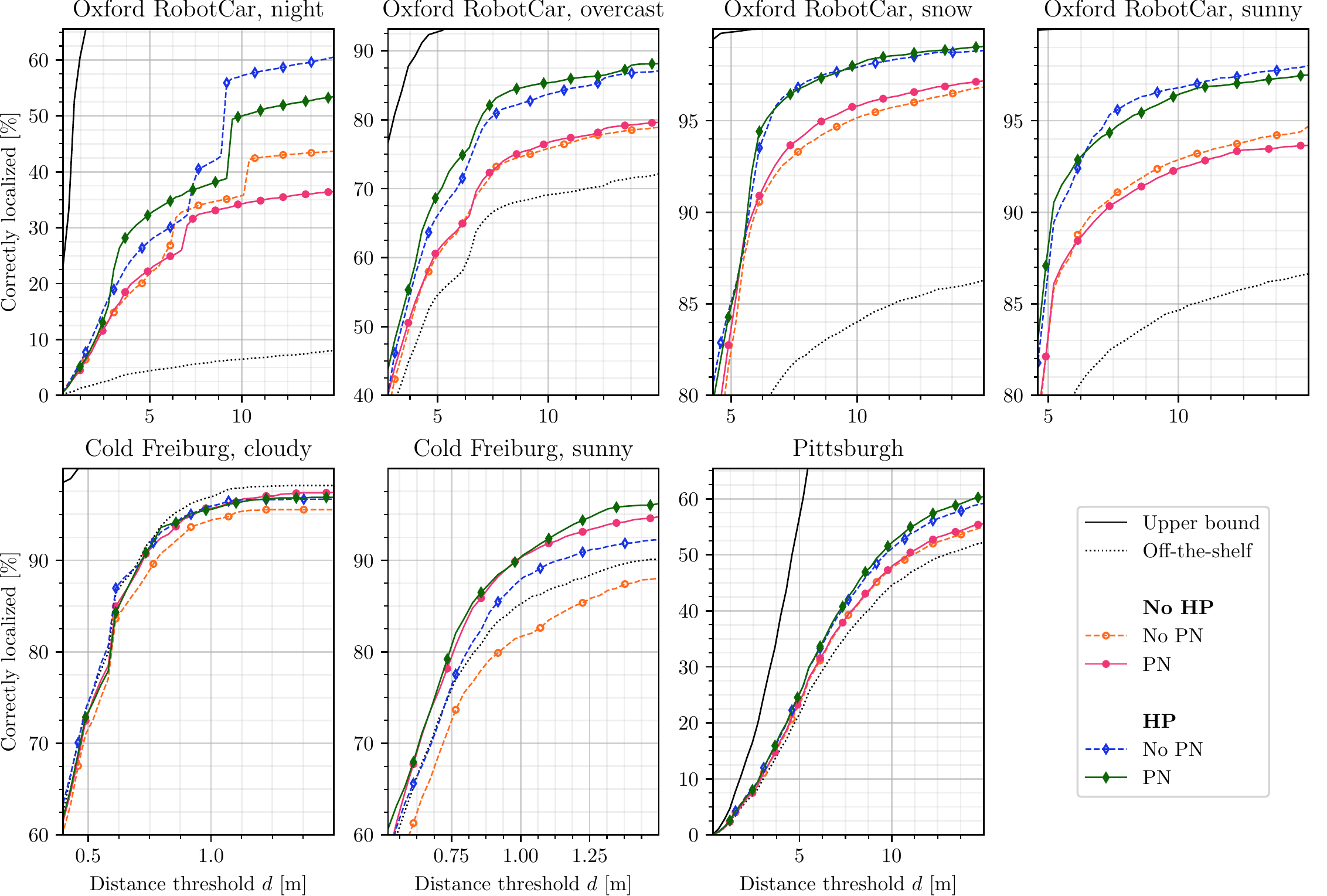}
    \caption{Localization accuracy ablation for hard positive (HP) and pairwise negative (PN) mining. All models are trained with triplet loss on large Oxford RobotCar (III).}
    \label{fig:mining}
\end{figure*}{}

\subsubsection{Hard Positive and Pairwise Negative Mining}
To evaluate the effects of hard positive (HP) and pairwise negative (PN) mining, we retrain one of our models while selectively switching off one or both of of our proposed mining strategies. Figure~\ref{fig:mining} shows the resulting localization accuracy for the seven different evaluation conditions. All models in Figure~\ref{fig:mining} are trained with triplet loss on large Oxford RobotCar (III). It can be observed that, for all but one condition (Cold Freiburg cloudy), hard positive mining is clearly beneficial for localization accuracy. Hard positive mining is used to learn invariances towards moving objects (e.g.\ cars) and severe light changes. Given that moving objects and light changes are rare in an indoor environment such as Cold Freiburg, it is not surprising that the boost in accuracy provided by hard positive mining is least significant for this dataset. 
When the models trained on Oxford RobotCar are deployed for localization on other datasets, such as Cold Freiburg or Pittsburgh, pairwise negative mining becomes crucial. This is because pairwise negative mining prevents over-fitting to particularly hard parts of the training set. If pairwise negative mining is dropped, the performance on the most difficult regions increases but the generalizability to new datasets is lost. An observant reader may notice that throughout our paper, where not stated otherwise, all models trained on large Oxford RobotCar (III), use pairwise negative mining.

\subsection{Additional Quantitative Results}
    \label{app:quantitative}
\begin{figure*}[htp]
    \centering
    \includegraphics[width=0.9\textwidth]{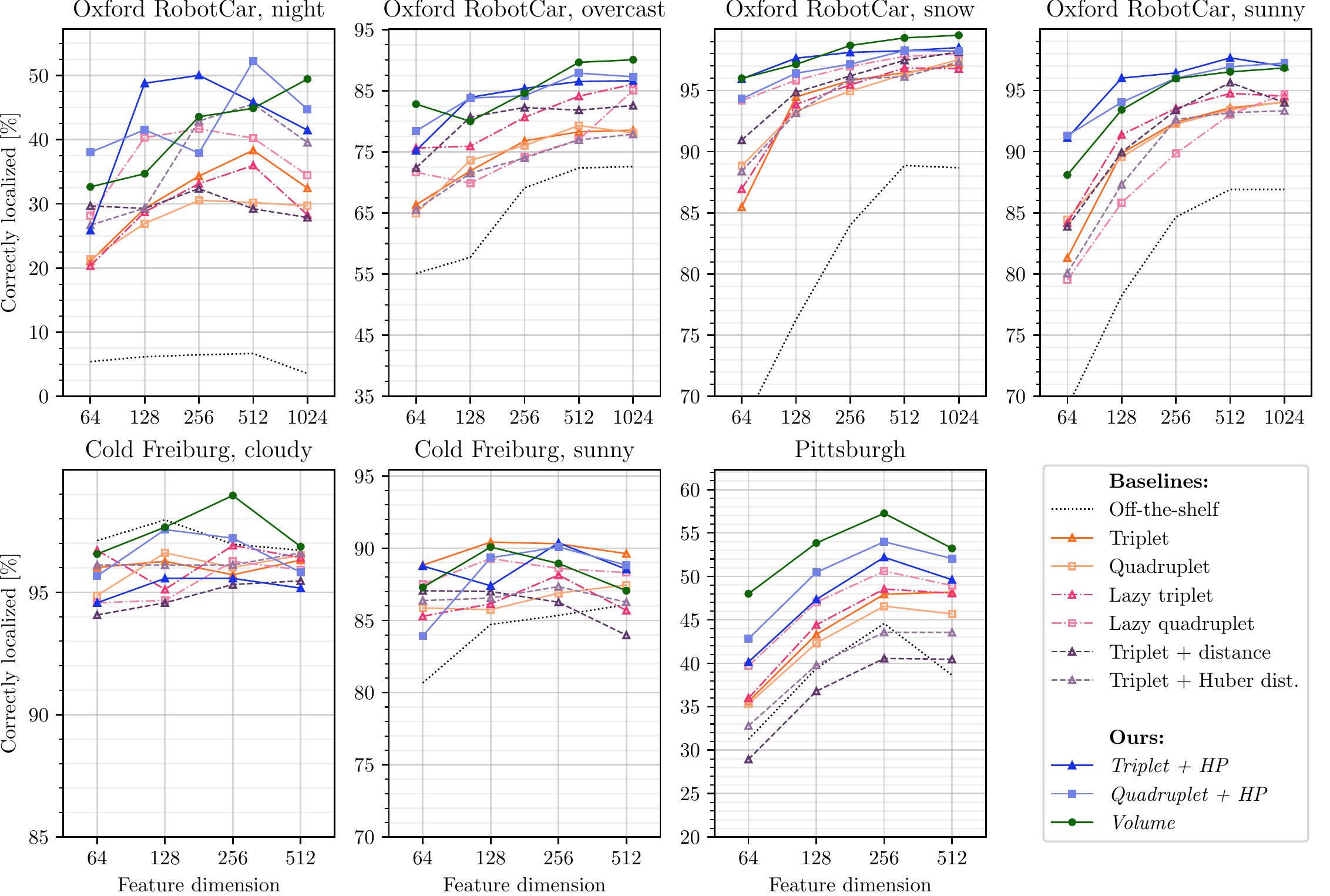}
    \caption{Localization accuracy SOTA comparison for models trained on large Oxford RobotCar (III) as a function of the feature dimension after PCA.}
    \label{fig:dims}
\end{figure*}

\begin{figure*}
    \centering
    \includegraphics[width=0.9\textwidth]{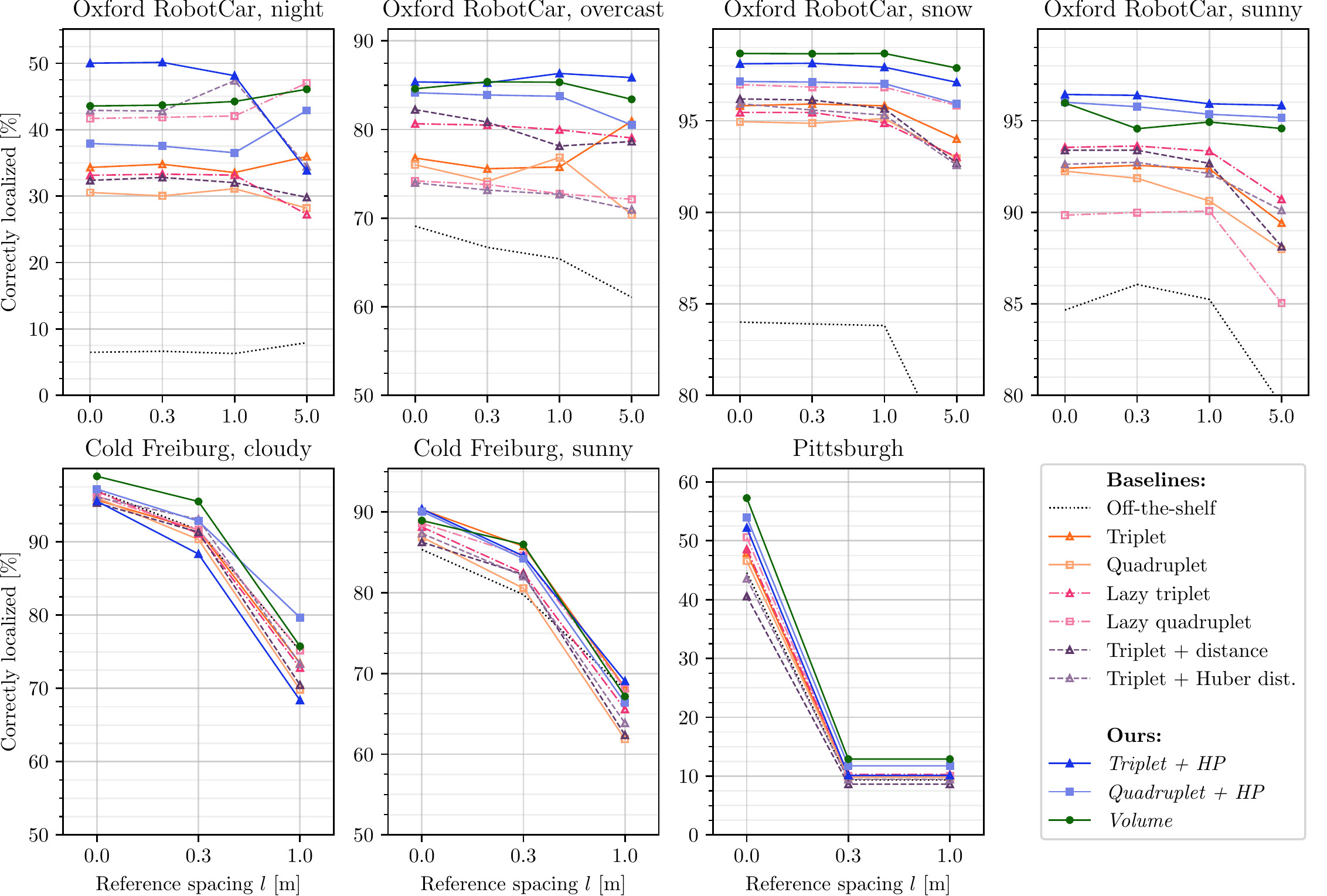}
    \caption{Localization accuracy SOTA comparison for models trained on large Oxford RobotCar (III) as a function of the distance $l$ between two consecutive reference images.}
    \label{fig:spacing}
\end{figure*}

\begin{figure*}
    \centering
    \includegraphics[width=0.9\textwidth]{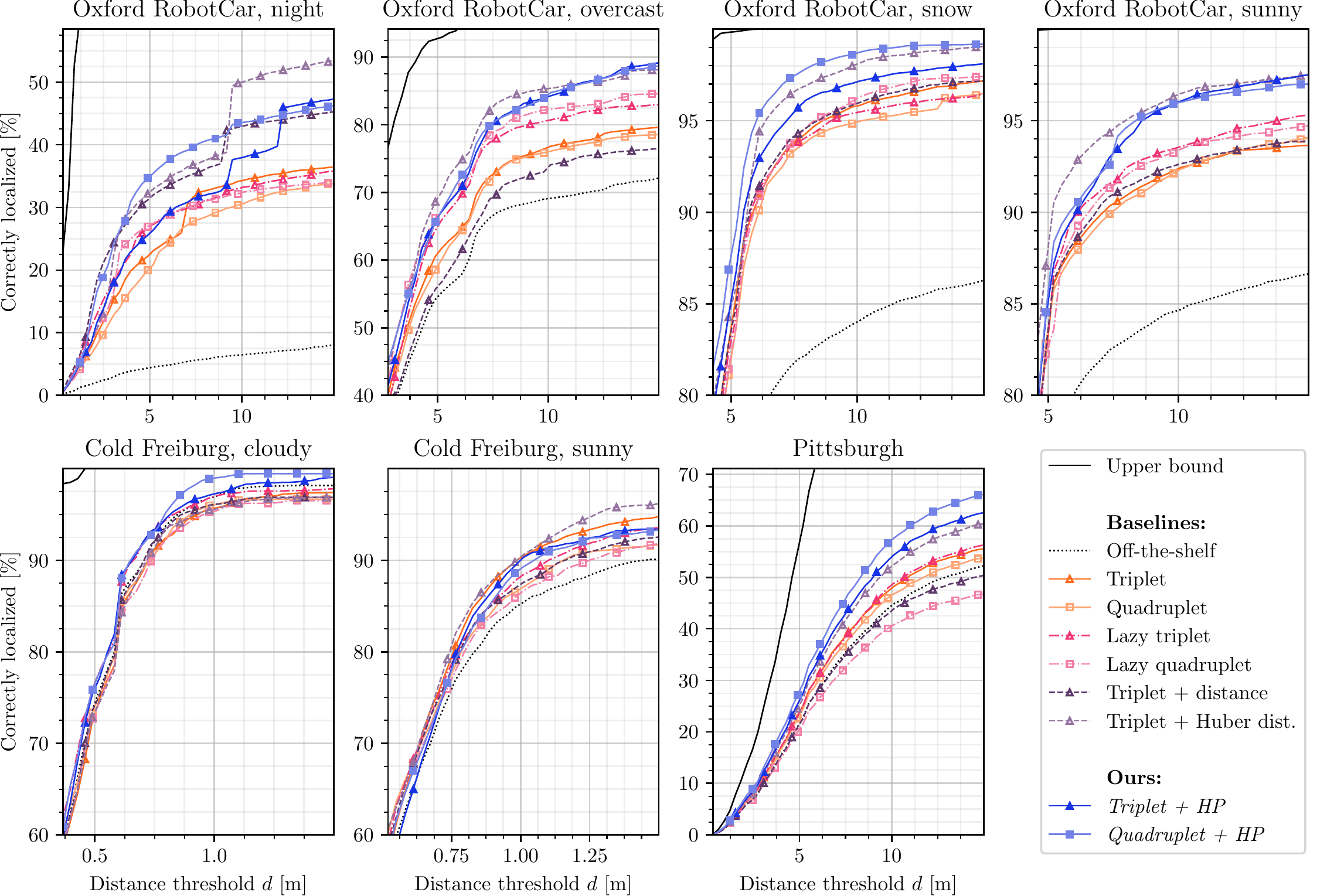}
    \caption{Localization accuracy SOTA comparison for models trained on large Oxford RobotCar (III) as a function of distance threshold $d$ for correct localization.}
    \label{fig:losses}
\end{figure*}

    Figure~\ref{fig:roc} shows a localization accuracy SOTA comparison for models trained on large Oxford RobotCar (III) as a function of feature dimension, reference spacing $l$ and distance threshold $d$. For brevity, only evaluations on Oxford RobotCar are included in the paper. Figures \ref{fig:dims}, \ref{fig:spacing}, and \ref{fig:losses} show the same comparison on all seven evaluation regions\footnote{Unless stated otherwise, we use $\text{dim}=256$, $l=0.0$, and $d=10.0$/$d=1.0$ outdoors/indoors throughout the entire paper.}. Due to the simple nature of our landmark selection algorithm, which does not take into account that for Cold Freiburg and Pittsburgh, there exist images with similar translation but large variations in rotation, all methods fail for larger $l$ on these two datasets. This problem could be addressed by using a better landmark sampling strategy such as~\cite{Thoma2019}. However, such an evaluation is not the focus of our paper. 

{}

{}

{}

\subsection{Feature Visualization}
\label{app:feature_visualizations}
To better understand the influence of training data, loss function and sampling, we visualize the features learned by the different setups in our paper using t-SNE and Grad-CAM representations. 

\begin{figure*}[htp]\centering
\resizebox{\textwidth}{!}{%
\begin{tabular}{ccc}
\toprule
Oxford RobotCar & Pittsburgh & Cold Freiburg\\ \midrule
\includegraphics[height=4cm]{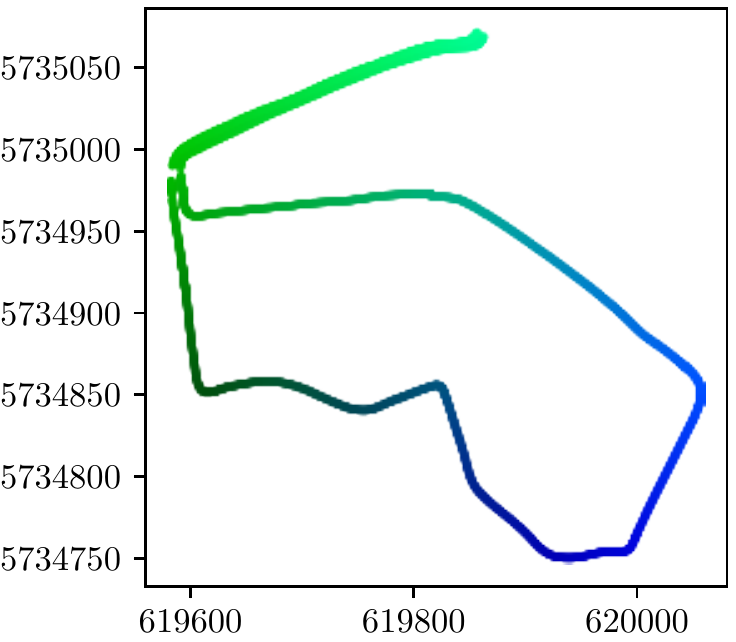}&
\includegraphics[height=4cm]{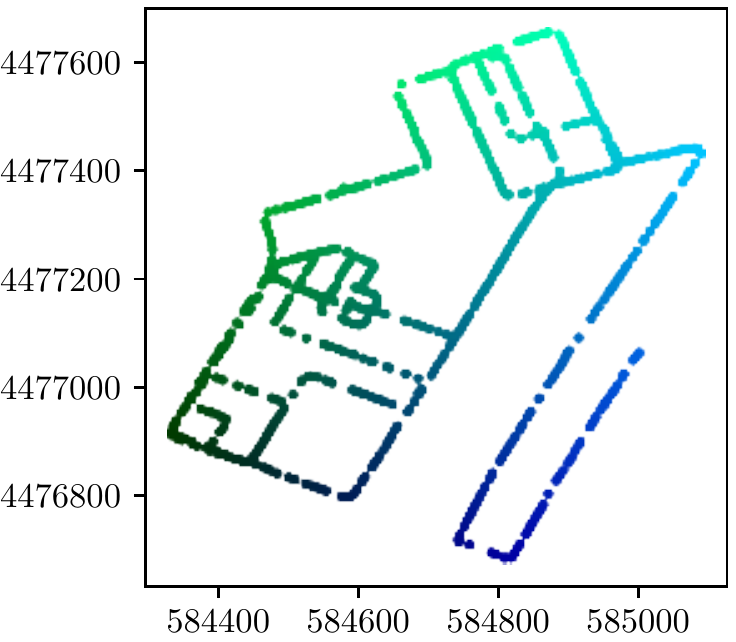}&
\includegraphics[height=4cm]{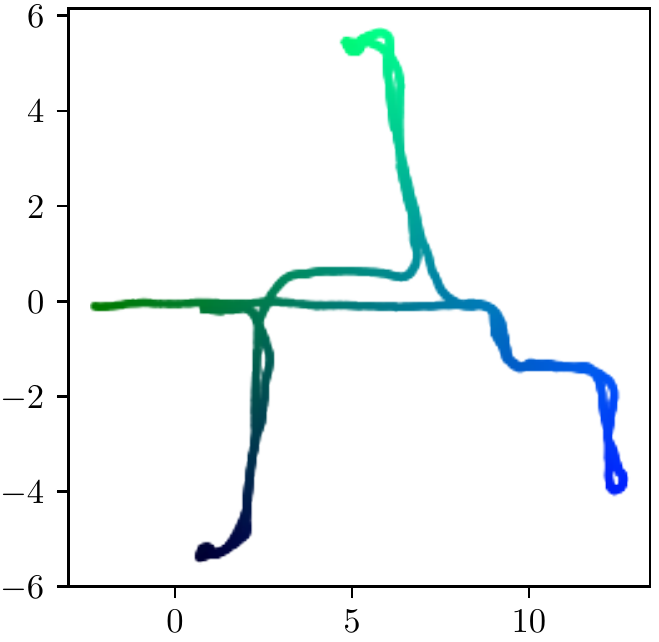}

\end{tabular}
}

\caption{Color code for t-SNE plots for the test regions from Oxford RobotCar, Pittsburgh and Cold Freiburg. Best viewed on screen.}
\label{fig:color_code}
\end{figure*}

\subsubsection{T-SNE}
\label{app:t-sne}
T-distributed Stochastic Neighbor Embedding (t-SNE) \cite{maaten2008visualizing} is a method for visualizing high dimensional data.
It defines two probability distributions, one for pairs of points in the original high-dimensional feature space and one for pairs of points in the low dimensional embedding. 
Pairs of similar points are given a high probability and dissimilar pairs a low probability. 
The points in the low dimensional map are then found by minimizing the Kullback-Leibler divergence between the two distributions.
This means that points that are located close in the embedding are---with high probability---also close in the original feature space, while points with high distance in the embedding are most likely also dissimilar in feature space.

We use color to encode the original locations. The color codes are shown in Figure~\ref{fig:color_code}. 
To reduce the number of plots, we only take into consideration the best network of each training region. 
Looking at Table~\ref{tab:quantitative}, it can be seen that those networks are:

\begin{enumerate}[I]
\item Quadruplet on Cold Freiburg
\item Triplet on Oxford RobotCar (small)
\item Volume with our mining on Oxford RobotCar (large)
\item Triplet on Pittsburgh30K
\item Initialization with ImageNet, no localization training
\end{enumerate}

\begin{figure*}[!t]
\begin{center}
\resizebox{\textwidth}{!}{%
\begin{tabular}{ccccc}
\toprule
&\multicolumn{4}{c}{Oxford RobotCar}\\
&Sunny & Overcast & Snow & Night\\ \midrule
\parbox{0.5cm}{I}&
\parbox{0.2\linewidth}{\includegraphics[width=\linewidth]{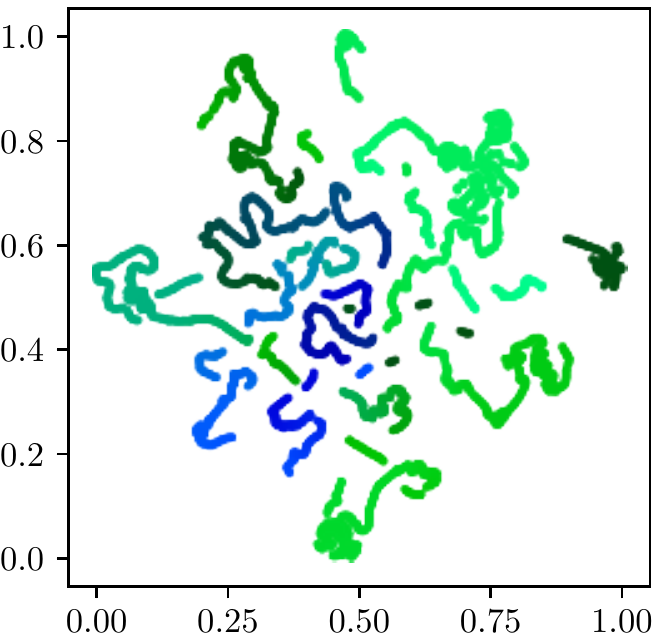}}&
\parbox{0.2\linewidth}{\includegraphics[width=\linewidth]{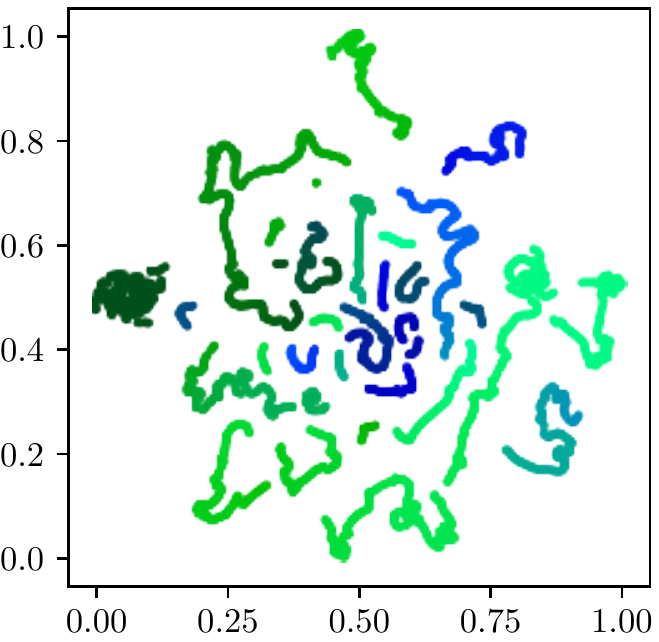}}&
\parbox{0.2\linewidth}{\includegraphics[width=\linewidth]{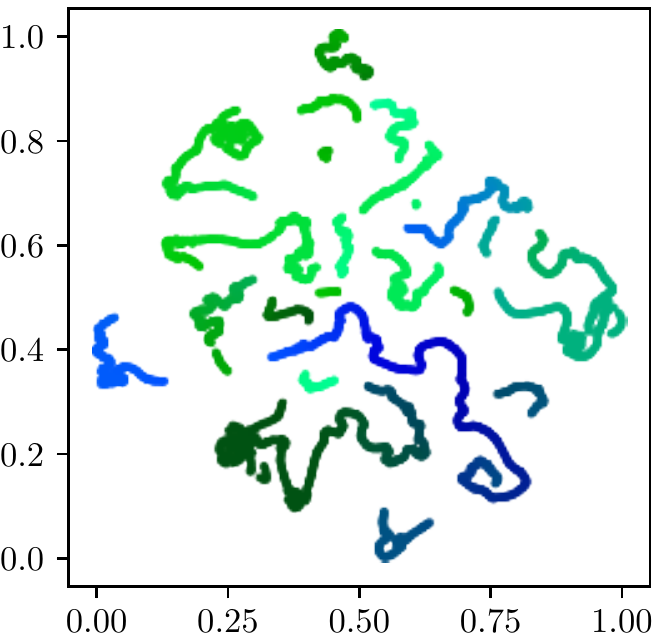}}&
\parbox{0.2\linewidth}{\includegraphics[width=\linewidth]{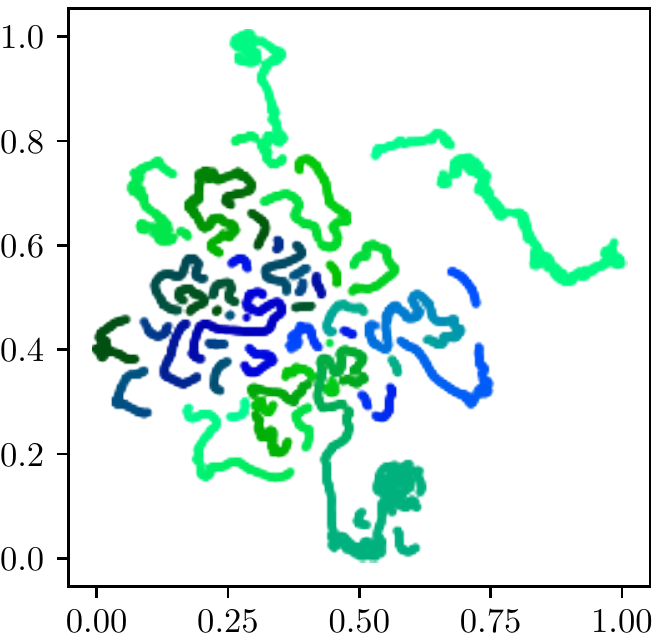}}\\

\parbox{0.5cm}{II}&
\parbox{0.2\linewidth}{\includegraphics[width=\linewidth]{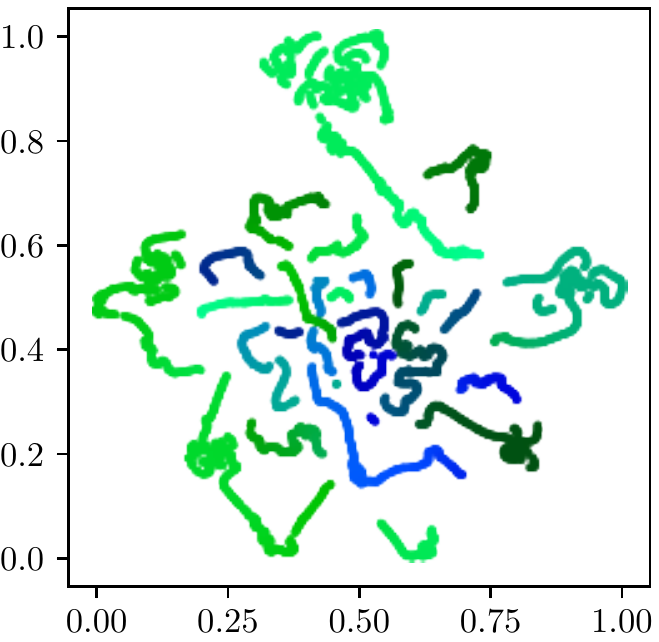}}&
\parbox{0.2\linewidth}{\includegraphics[width=\linewidth]{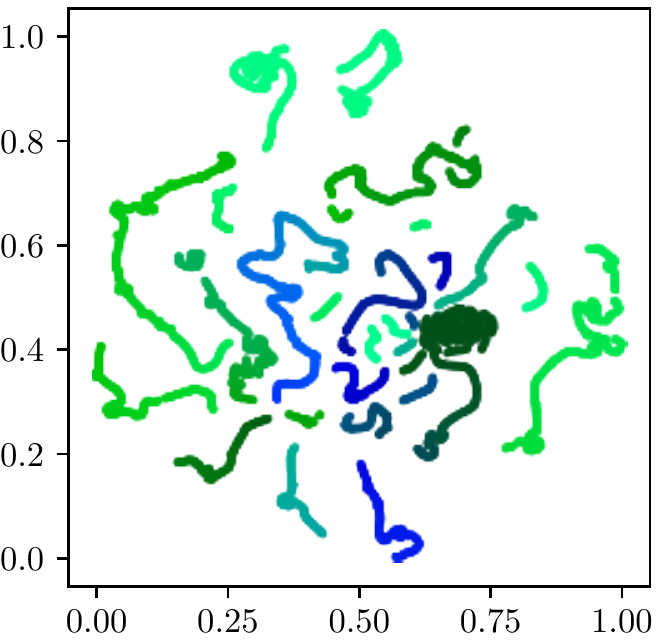}}&
\parbox{0.2\linewidth}{\includegraphics[width=\linewidth]{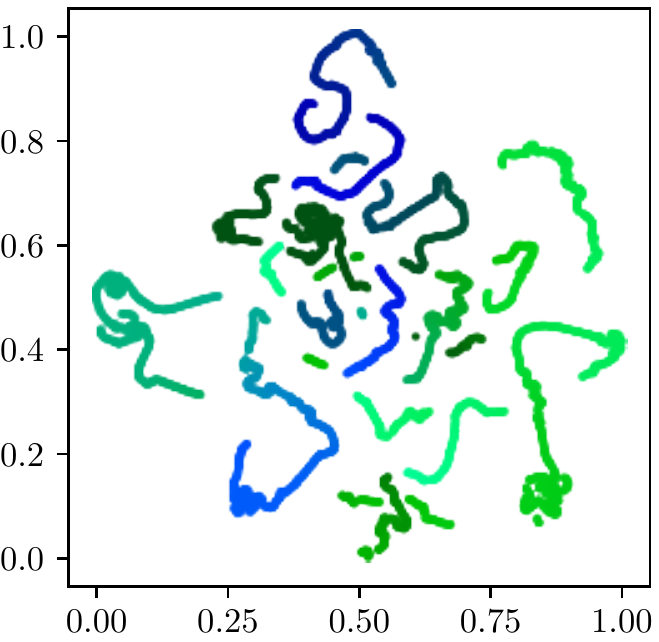}}&
\parbox{0.2\linewidth}{\includegraphics[width=\linewidth]{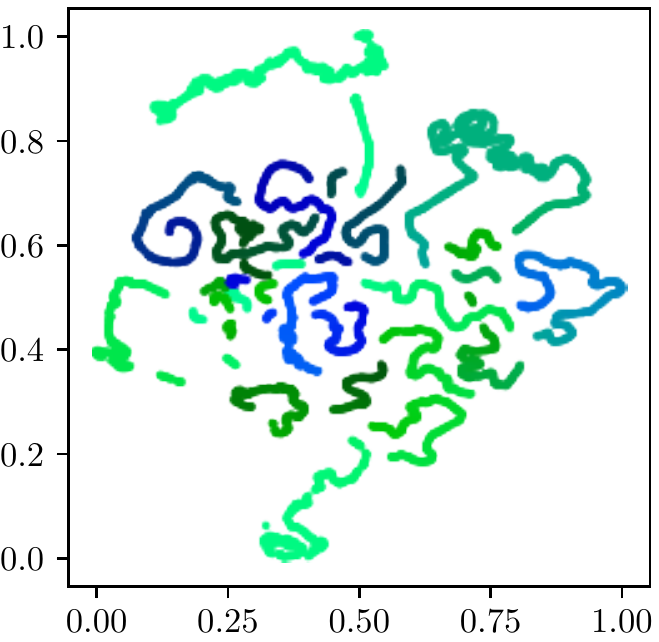}}\\

\parbox{0.5cm}{III}&
\parbox{0.2\linewidth}{\includegraphics[width=\linewidth]{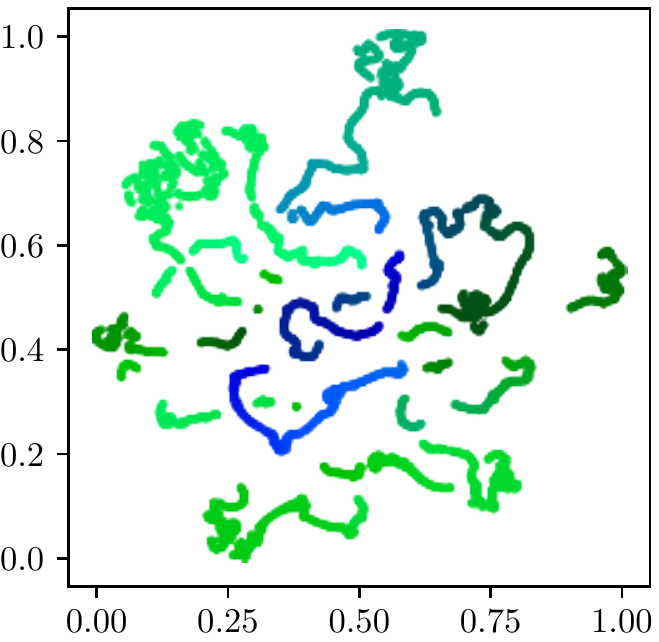}}&
\parbox{0.2\linewidth}{\includegraphics[width=\linewidth]{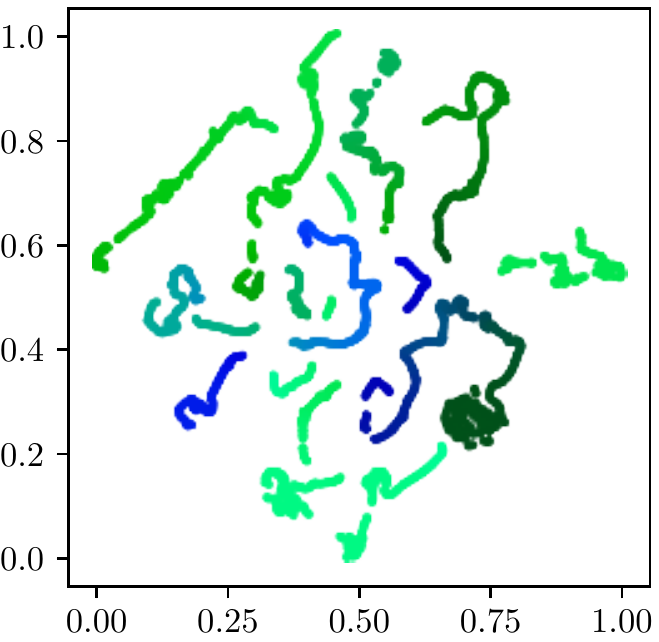}}&
\parbox{0.2\linewidth}{\includegraphics[width=\linewidth]{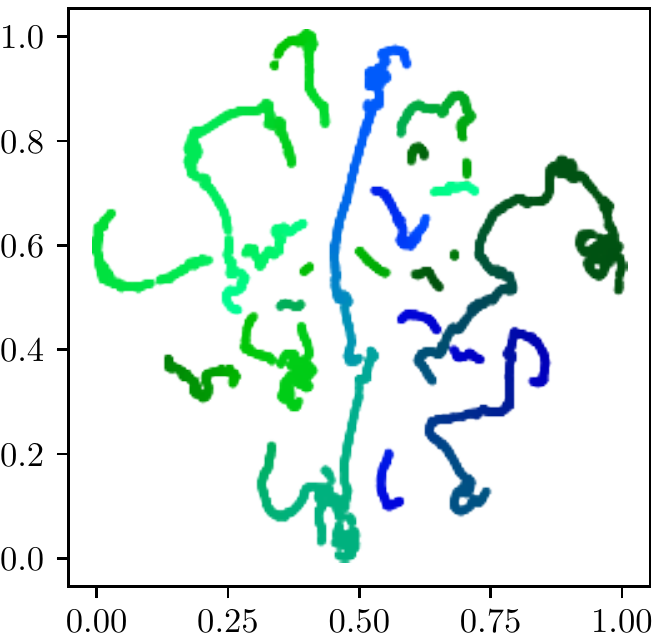}}&
\parbox{0.2\linewidth}{\includegraphics[width=\linewidth]{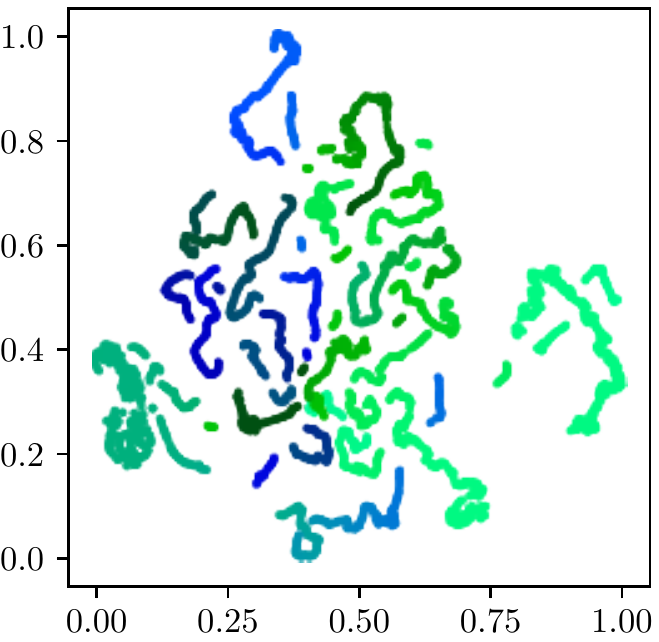}}\\

\parbox{0.5cm}{IV}&
\parbox{0.2\linewidth}{\includegraphics[width=\linewidth]{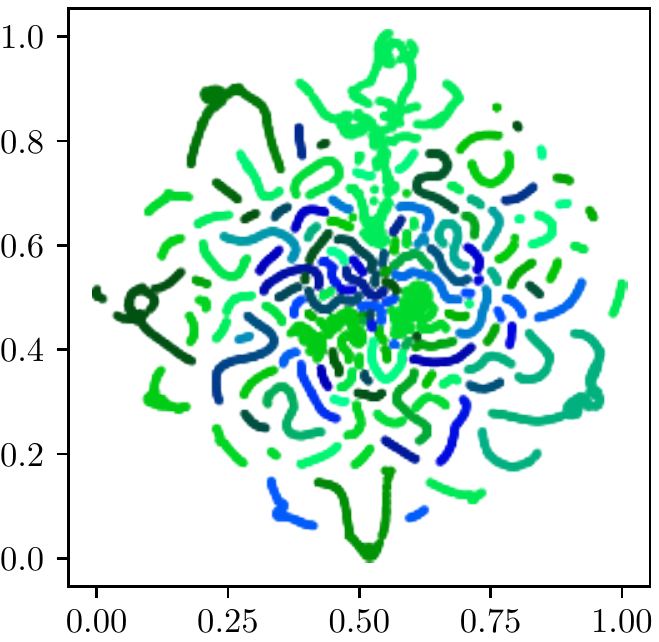}}&
\parbox{0.2\linewidth}{\includegraphics[width=\linewidth]{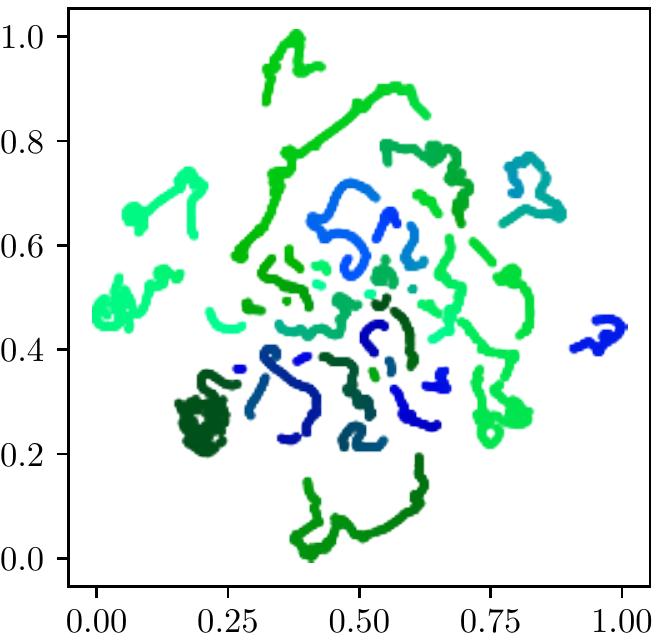}}&
\parbox{0.2\linewidth}{\includegraphics[width=\linewidth]{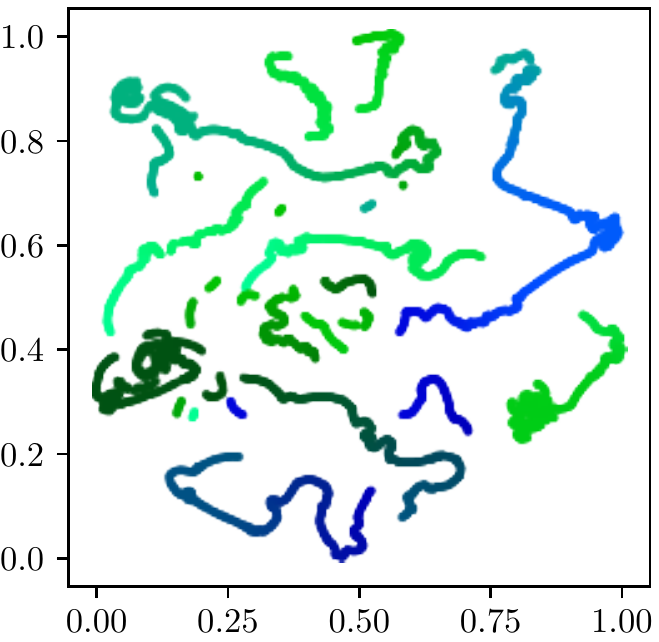}}&
\parbox{0.2\linewidth}{\includegraphics[width=\linewidth]{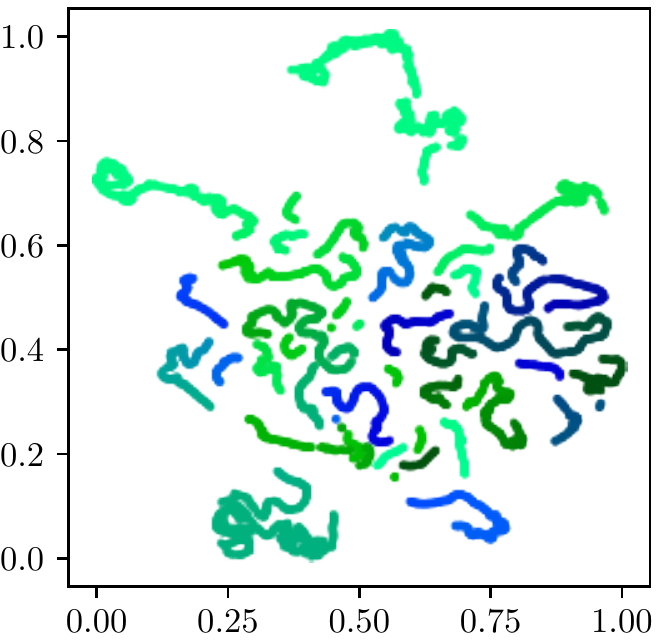}}\\

\parbox{0.5cm}{V}&
\parbox{0.2\linewidth}{\includegraphics[width=\linewidth]{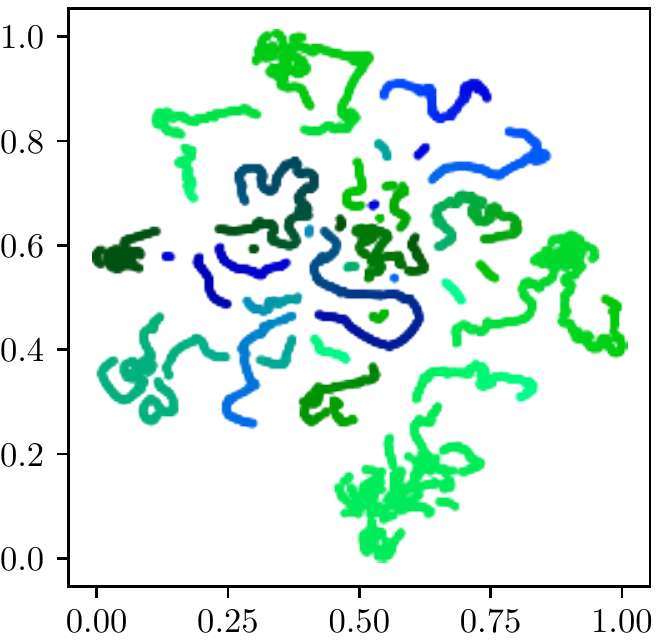}}&
\parbox{0.2\linewidth}{\includegraphics[width=\linewidth]{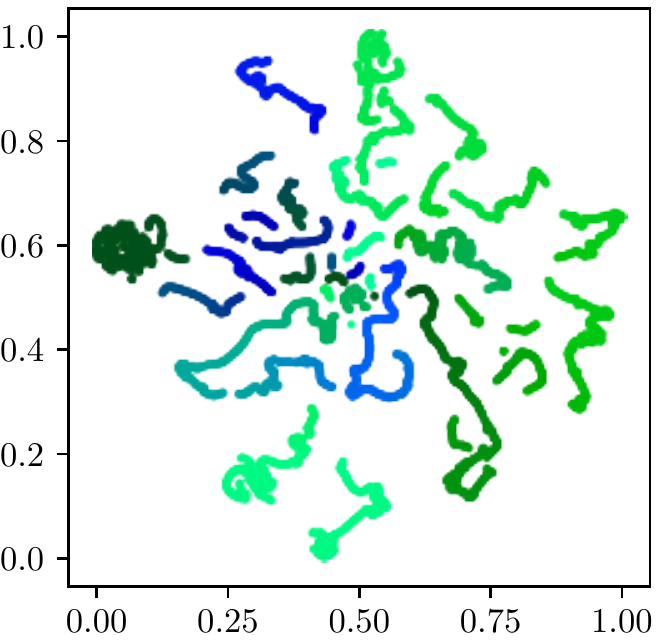}}&
\parbox{0.2\linewidth}{\includegraphics[width=\linewidth]{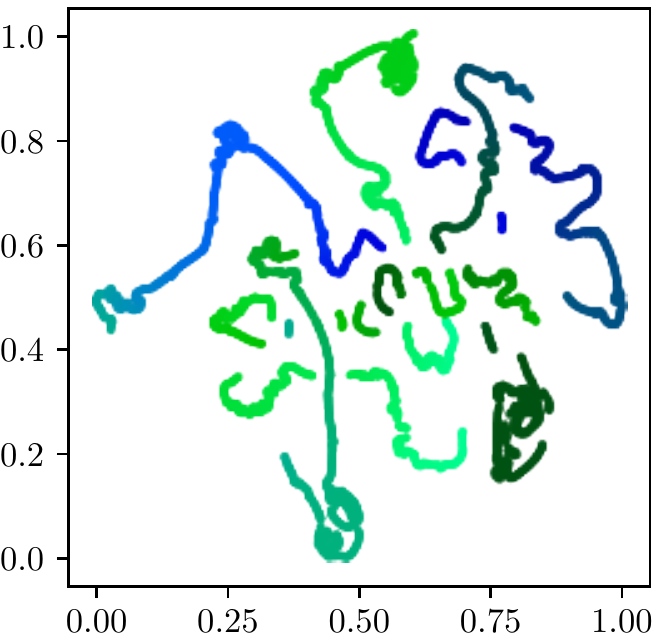}}&
\parbox{0.2\linewidth}{\includegraphics[width=\linewidth]{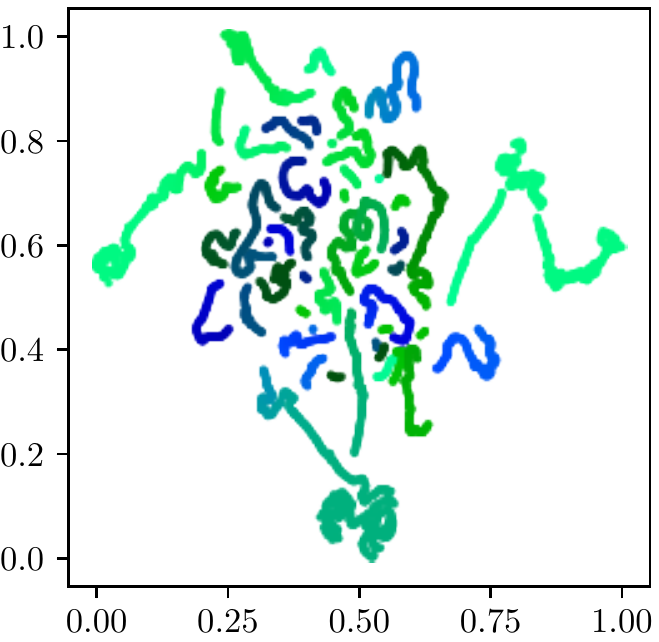}}\\

\end{tabular}
}
\vspace{-0.3cm}
\end{center}
\caption{T-SNE visualization of the query image feature distribution for different test regions inferred with the best network of each training region---I: Quadruplet on Cold Freiburg, II: Triplet on Oxford RobotCar (small), III: Volume with our mining on Oxford RobotCar (large), IV: Triplet on Pittsburgh30K, V: Initialization with ImageNet, no localization training. The colors correspond to different locations as shown in Figure~\ref{fig:color_code}. Best viewed on screen.}
\label{fig:tsne_1}
\vspace{-0.2cm}
\end{figure*}

\begin{figure*}[p]
\begin{center}
\resizebox{0.8\textwidth}{!}{%
\begin{tabular}{cccc}
\toprule
&\multicolumn{2}{c}{Cold Freiburg} & Pittsburgh\\
&Sunny & Cloudy  & \\ \midrule
\parbox{0.5cm}{I}&
\parbox{0.3\linewidth}{\includegraphics[width=\linewidth]{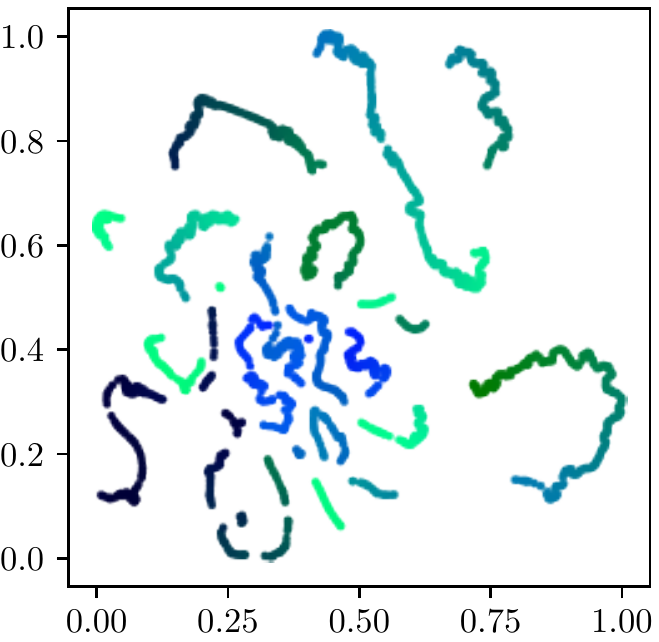}}&
\parbox{0.3\linewidth}{\includegraphics[width=\linewidth]{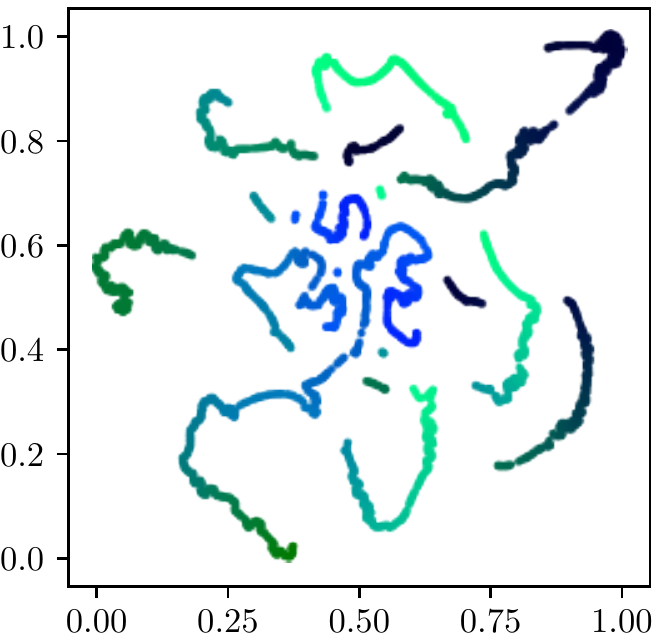}}&
\parbox{0.3\linewidth}{\includegraphics[width=\linewidth]{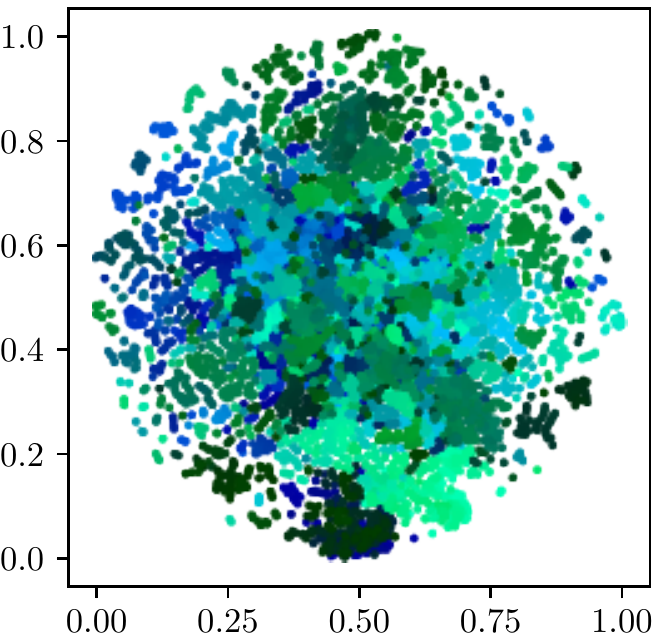}}\\

\parbox{0.5cm}{II}&
\parbox{0.3\linewidth}{\includegraphics[width=\linewidth]{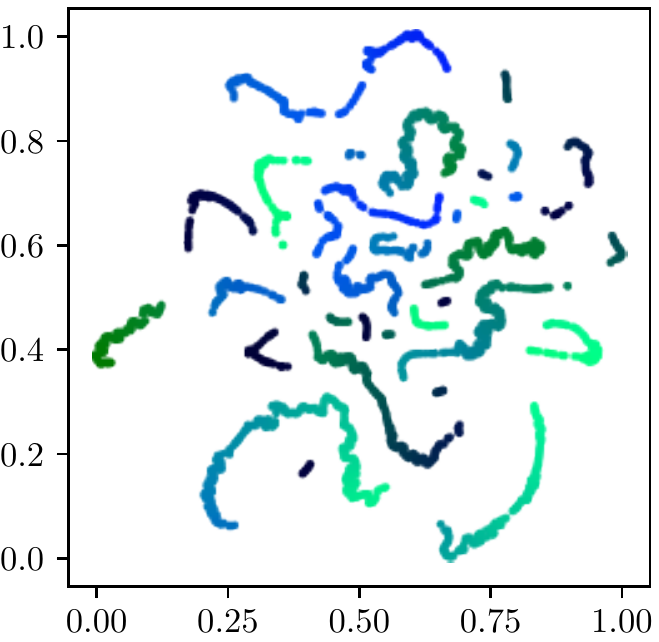}}&
\parbox{0.3\linewidth}{\includegraphics[width=\linewidth]{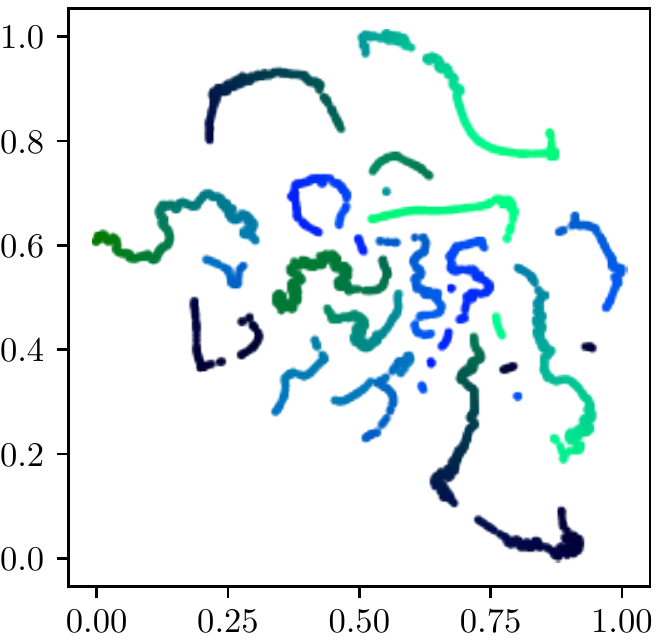}}&
\parbox{0.3\linewidth}{\includegraphics[width=\linewidth]{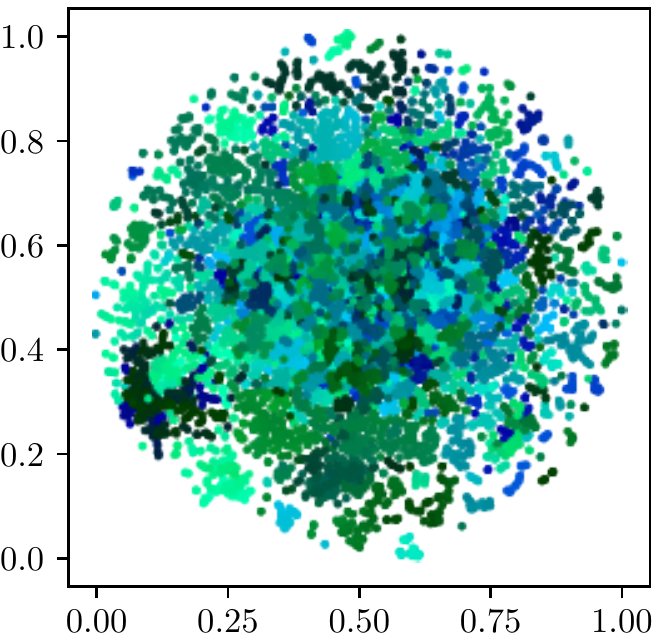}}\\

\parbox{0.5cm}{III}&
\parbox{0.3\linewidth}{\includegraphics[width=\linewidth]{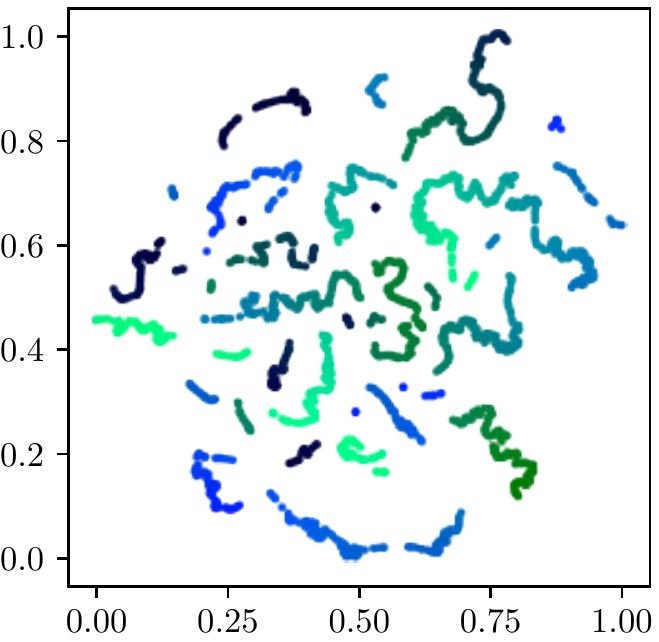}}&
\parbox{0.3\linewidth}{\includegraphics[width=\linewidth]{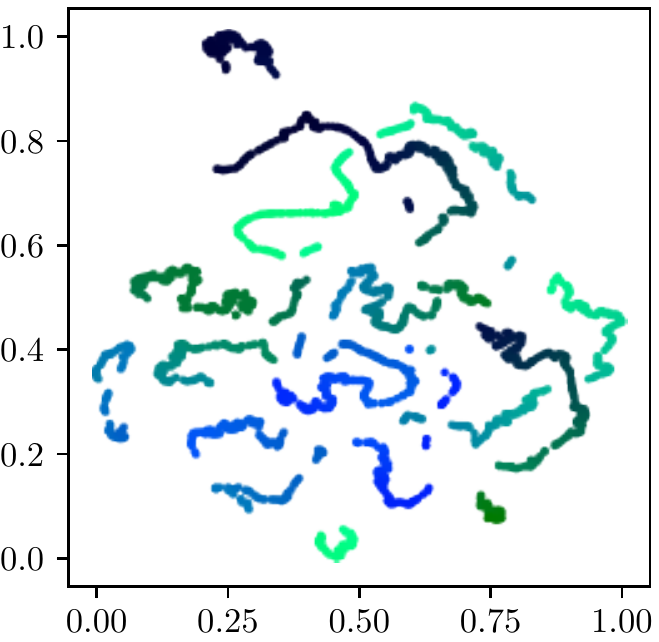}}&
\parbox{0.3\linewidth}{\includegraphics[width=\linewidth]{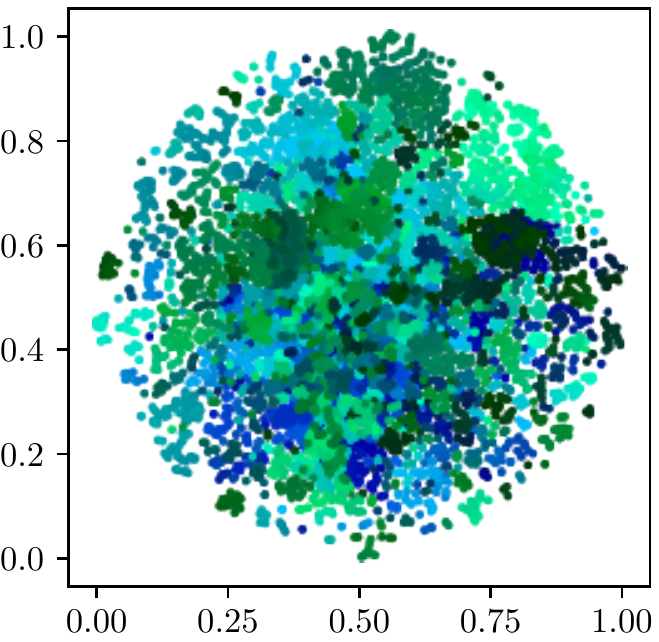}}\\

\parbox{0.5cm}{IV}&
\parbox{0.3\linewidth}{\includegraphics[width=\linewidth]{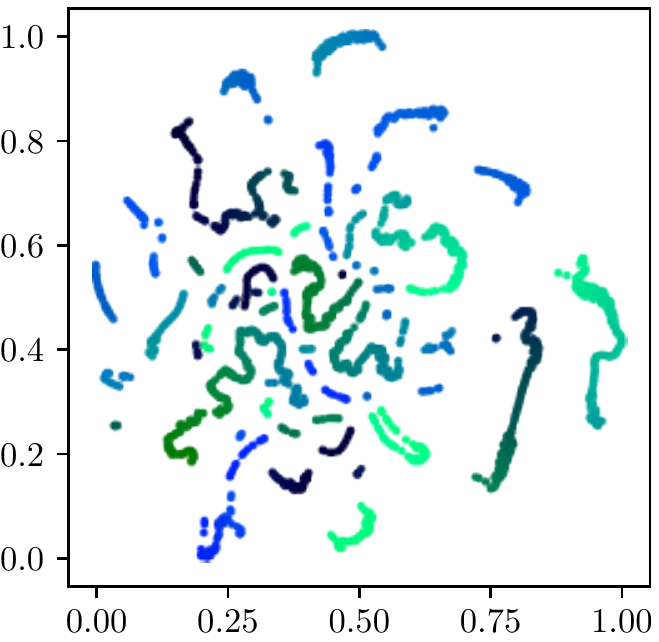}}&
\parbox{0.3\linewidth}{\includegraphics[width=\linewidth]{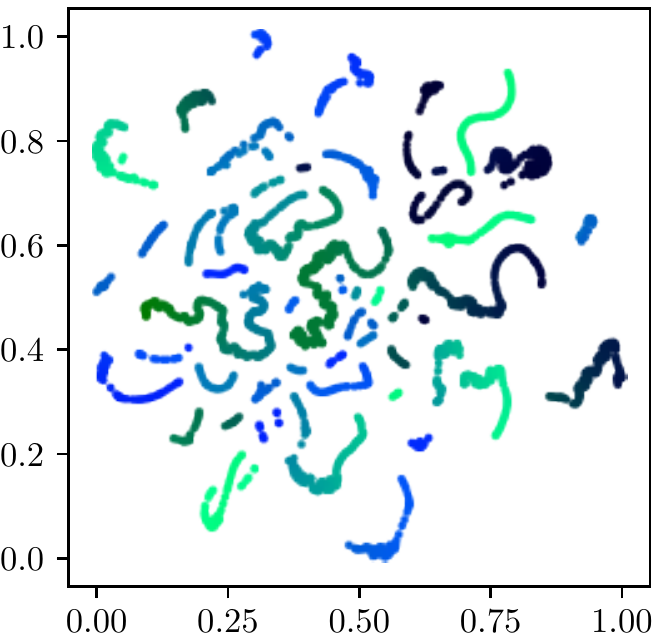}}&
\parbox{0.3\linewidth}{\includegraphics[width=\linewidth]{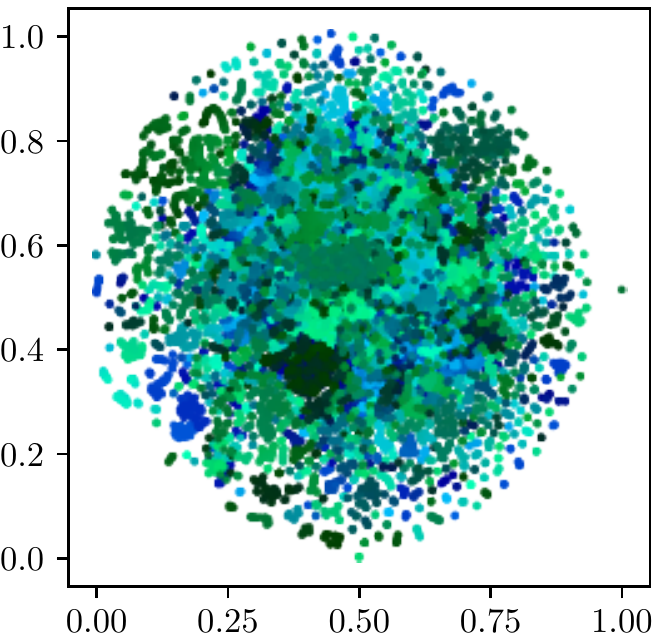}}\\

\parbox{0.5cm}{V}&
\parbox{0.3\linewidth}{\includegraphics[width=\linewidth]{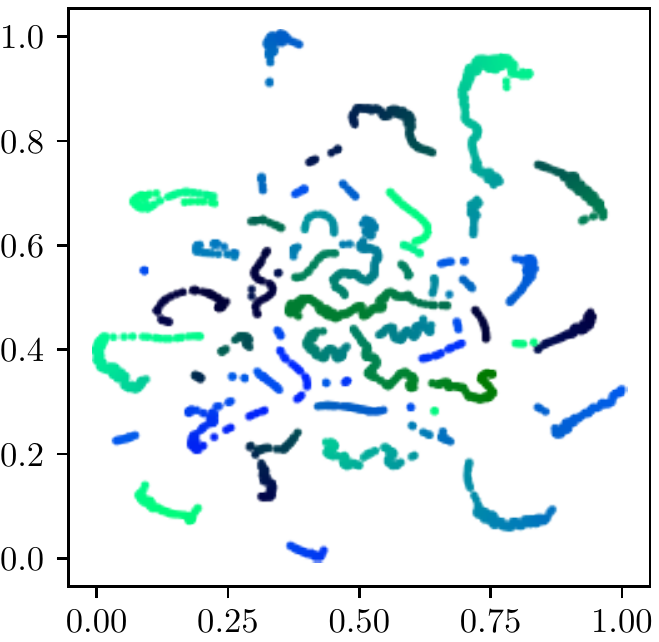}}&
\parbox{0.3\linewidth}{\includegraphics[width=\linewidth]{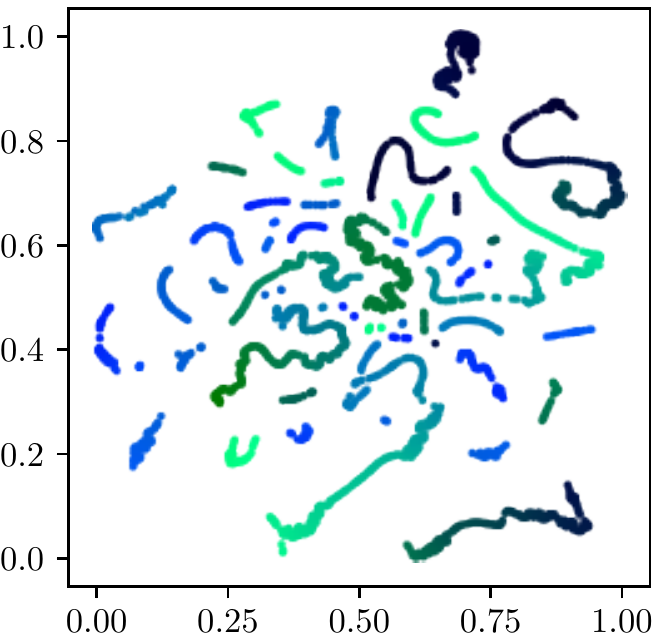}}&
\parbox{0.3\linewidth}{\includegraphics[width=\linewidth]{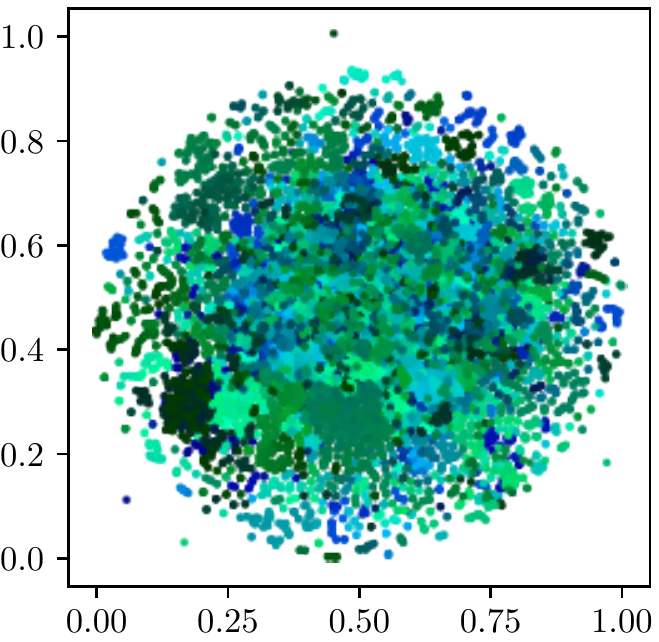}}\\
\end{tabular}
}
\end{center}
\caption{T-SNE visualization of the query image feature distribution for different test regions inferred with the best network of each training region---I: Quadruplet on Cold Freiburg, II: Triplet on Oxford RobotCar (small), III: Volume with our mining on Oxford RobotCar (large), IV: Triplet on Pittsburgh30K, V: Initialization with ImageNet, no localization training.  The colors correspond to different locations as shown in Figure~\ref{fig:color_code}. Best viewed on screen.}
\label{fig:tsne_2}
\end{figure*}

Figure~\ref{fig:tsne_1} shows the t-SNE visualizations for the Oxford RobotCar test regions while Figure~\ref{fig:tsne_2} shows the t-SNE visualizations for Cold Freiburg and Pittsburgh. 
Ideally, features that are close in geometric space, should also be close in feature space.
This means that the t-SNE plot of a good feature should cluster similar colors together. Looking at Figures~\ref{fig:tsne_1} and \ref{fig:tsne_2} reveals that this is most prominently the case for features trained with our volume-based loss on the large Oxford RobotCar (III) training region.
This finding aligns with the quantitative evaluation in Table 2, which also indicates that---in the mean over all seven evaluation regions---our volume-based loss with our mining performs best. 

\subsubsection{Grad-CAM}
    \label{app:grad-cam}
\begin{figure*}[p]
\begin{center}
\resizebox{0.84\textwidth}{!}{%
\begin{tabular}{cc}
\toprule
\multicolumn{2}{c}{Oxford RobotCar, Sunny}\\ \hline
\parbox{0.5\linewidth}{\includegraphics[width=\linewidth]{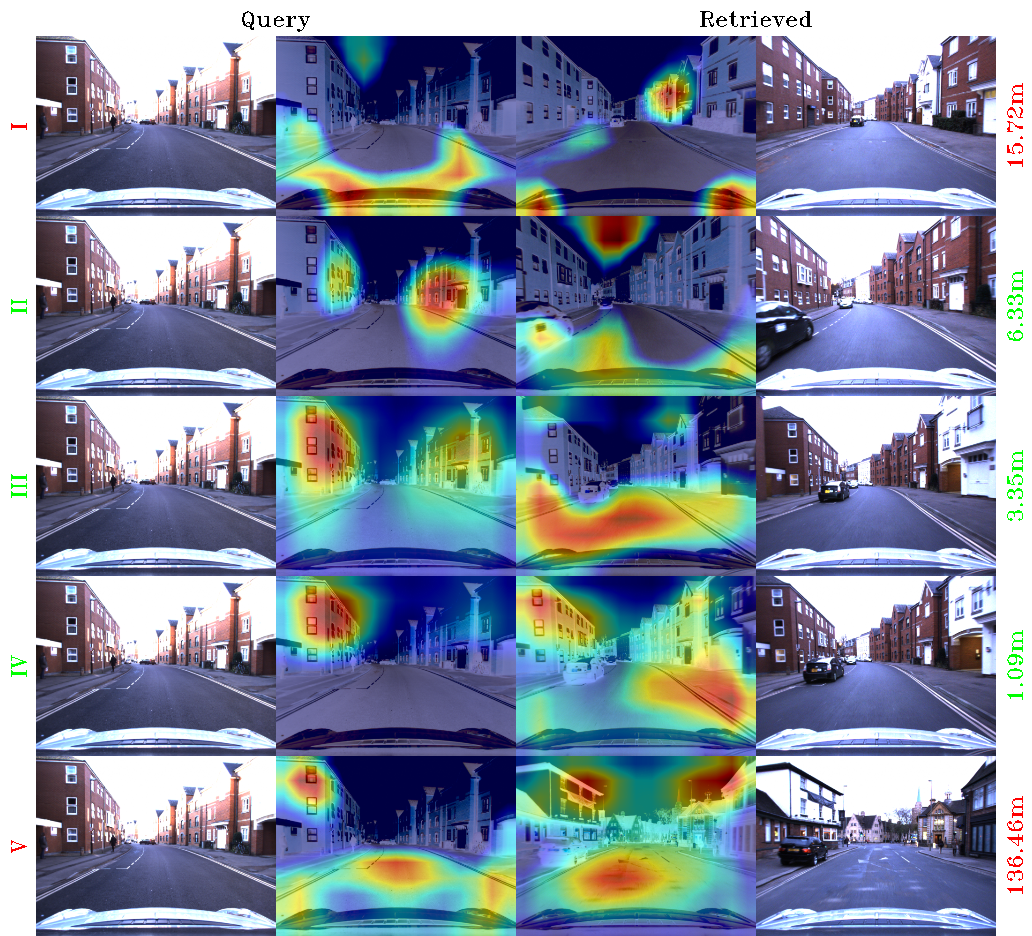}}&
\parbox{0.5\linewidth}{\includegraphics[width=\linewidth]{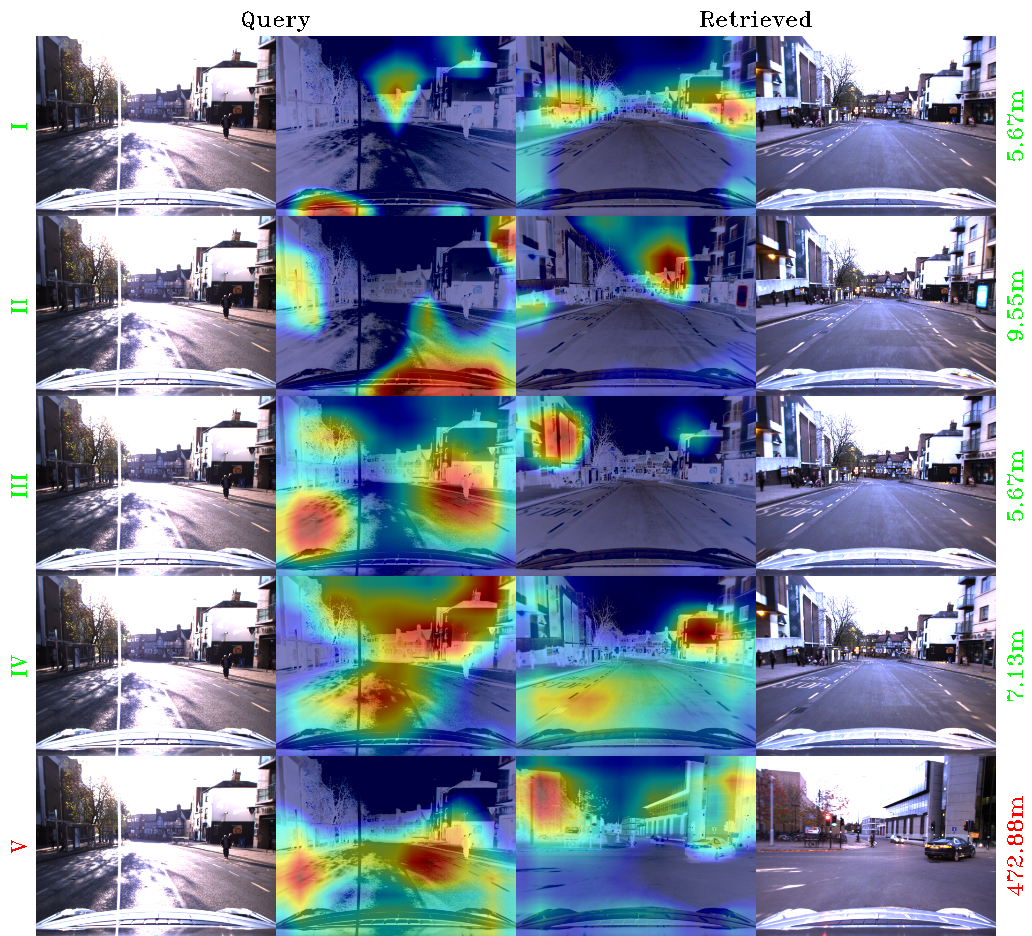}}\\
\\
\toprule
\multicolumn{2}{c}{Oxford RobotCar, Overcast}\\ \hline
\parbox{0.5\linewidth}{\includegraphics[width=\linewidth]{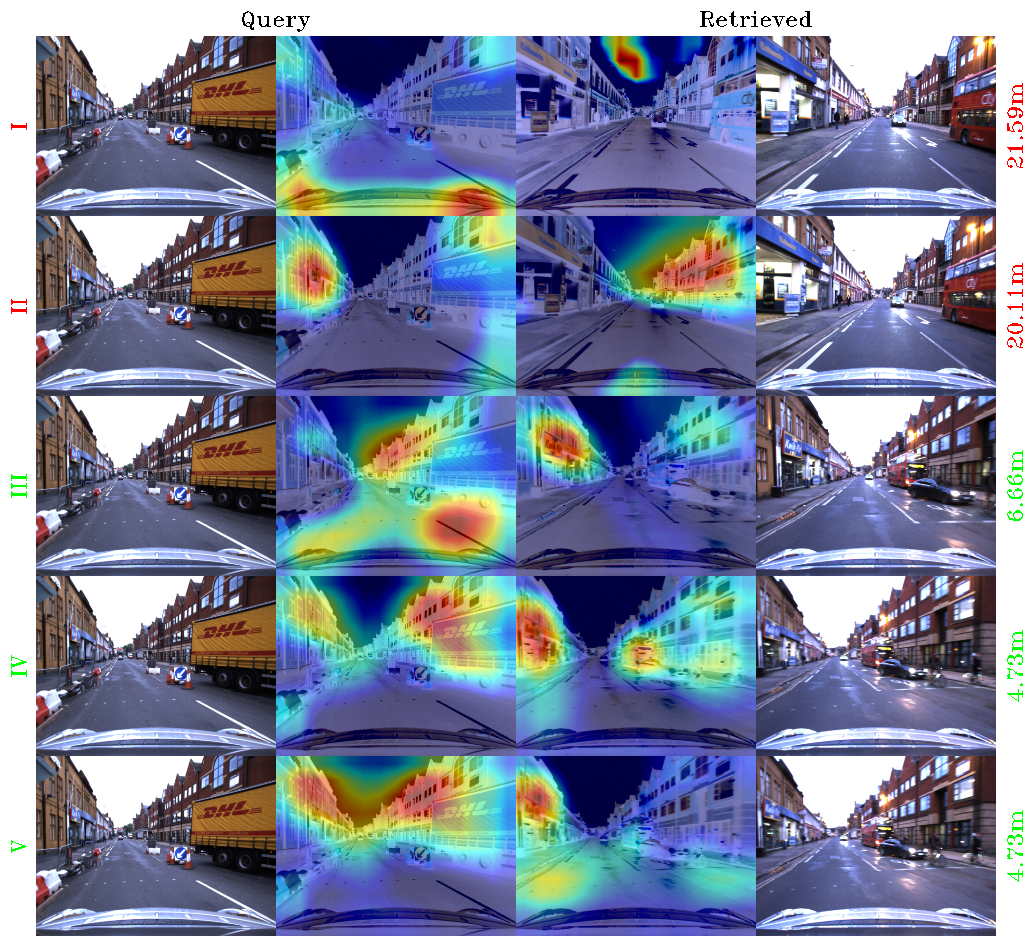}}&
\parbox{0.5\linewidth}{\includegraphics[width=\linewidth]{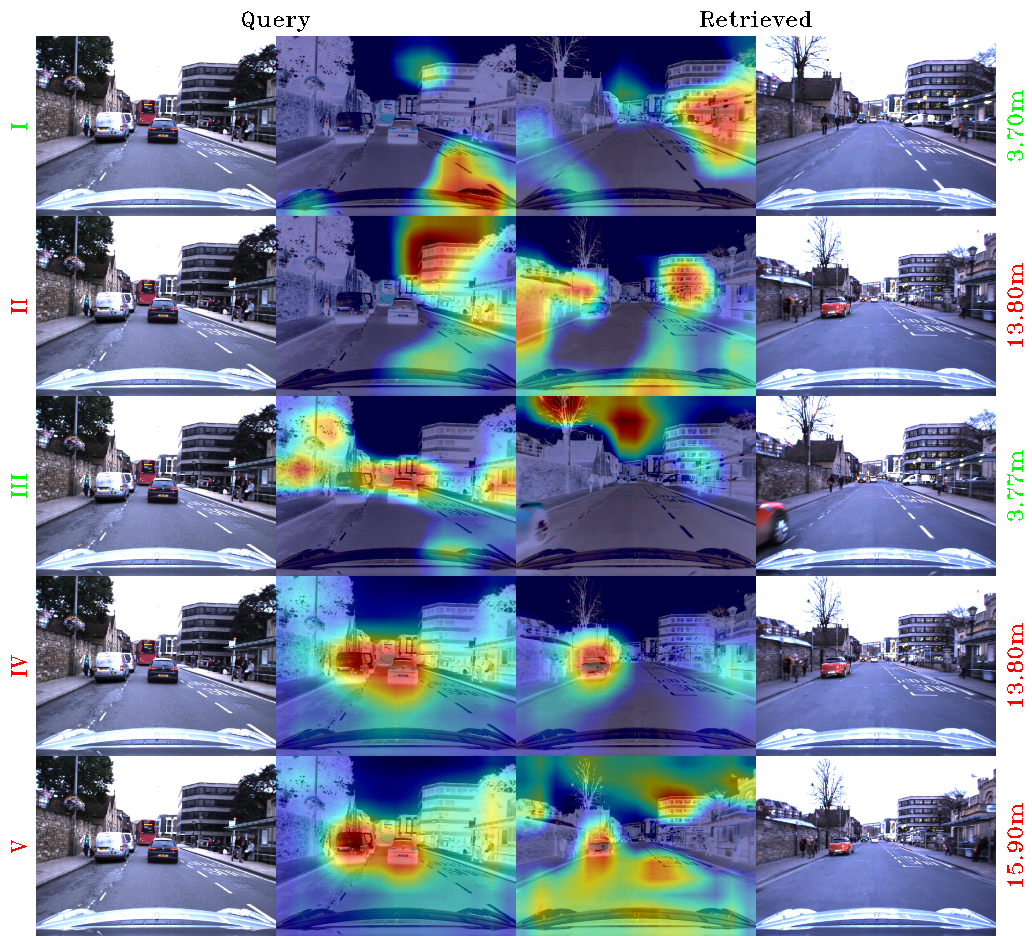}}\\
\\
\toprule
\multicolumn{2}{c}{Oxford RobotCar, Snow}\\ \hline
\parbox{0.5\linewidth}{\includegraphics[width=\linewidth]{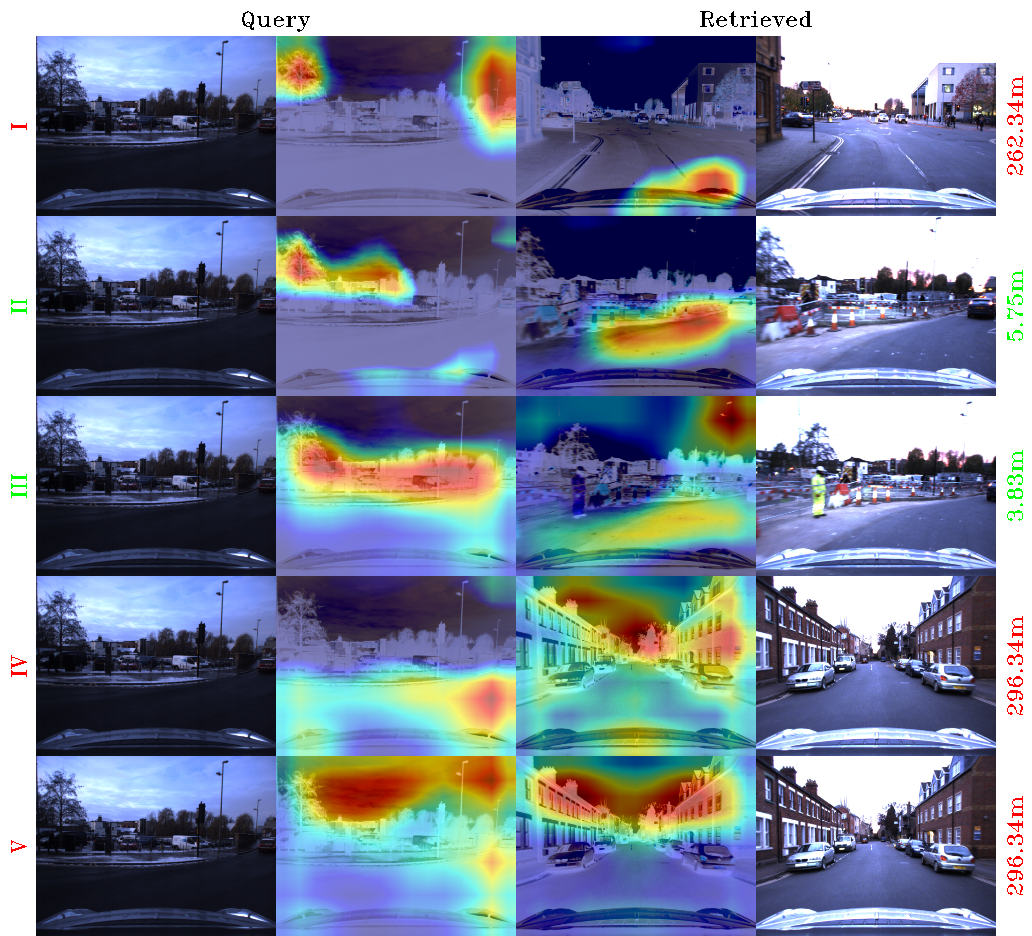}}&
\parbox{0.5\linewidth}{\includegraphics[width=\linewidth]{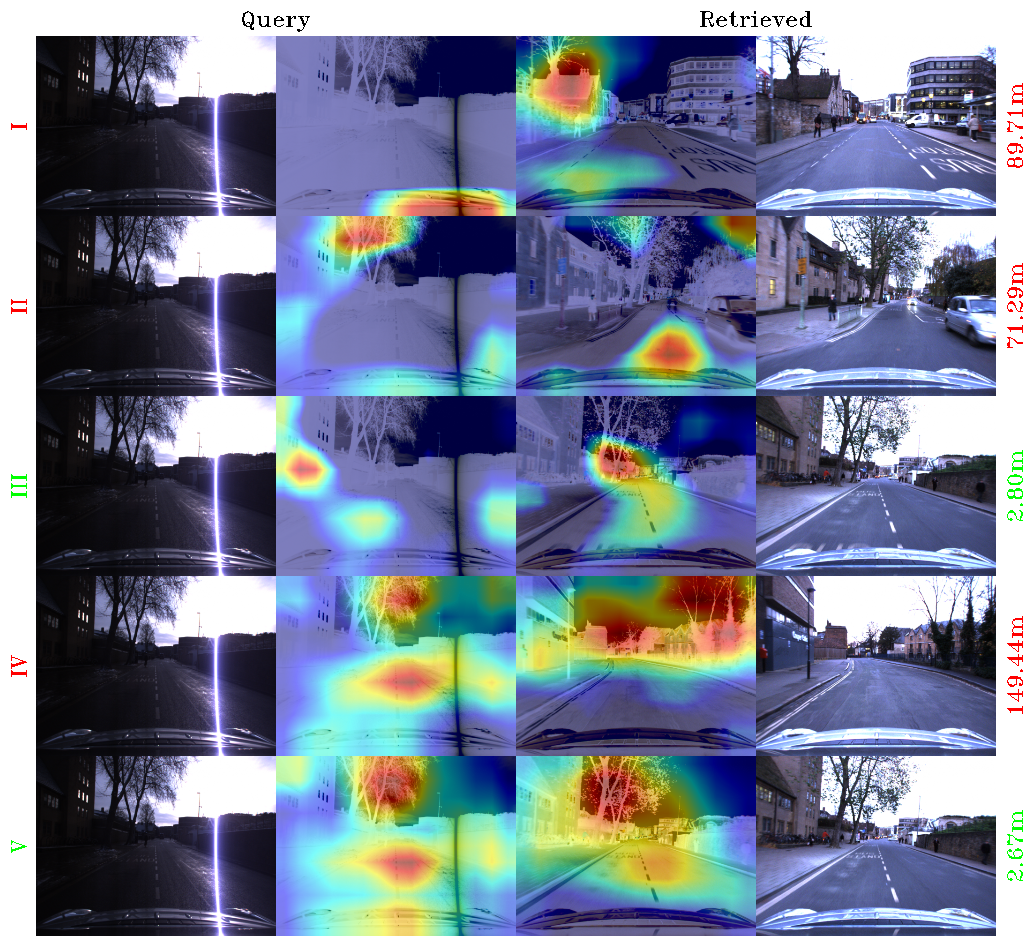}}\\
\end{tabular}
}
\end{center}
\caption{Selected visual results from the best network of each training region---I: Quadruplet on Cold Freiburg, II: Triplet on Oxford RobotCar (small), III: Volume with our mining on Oxford RobotCar (large), IV: Triplet on Pittsburgh30K, V: Initialization with ImageNet, no localization training. For each example we show the query image, Grad-CAM on the query image, Grad-CAM on the retrieved image, retrieved image, and the distance between retrieved image and query image.}
\label{fig:visuals_1}
\end{figure*}

\begin{figure*}[p]
\begin{center}
\resizebox{0.84\textwidth}{!}{%
\begin{tabular}{cc}
\toprule
\multicolumn{2}{c}{Oxford RobotCar, Night}\\ \hline
\parbox{0.5\linewidth}{\includegraphics[width=\linewidth]{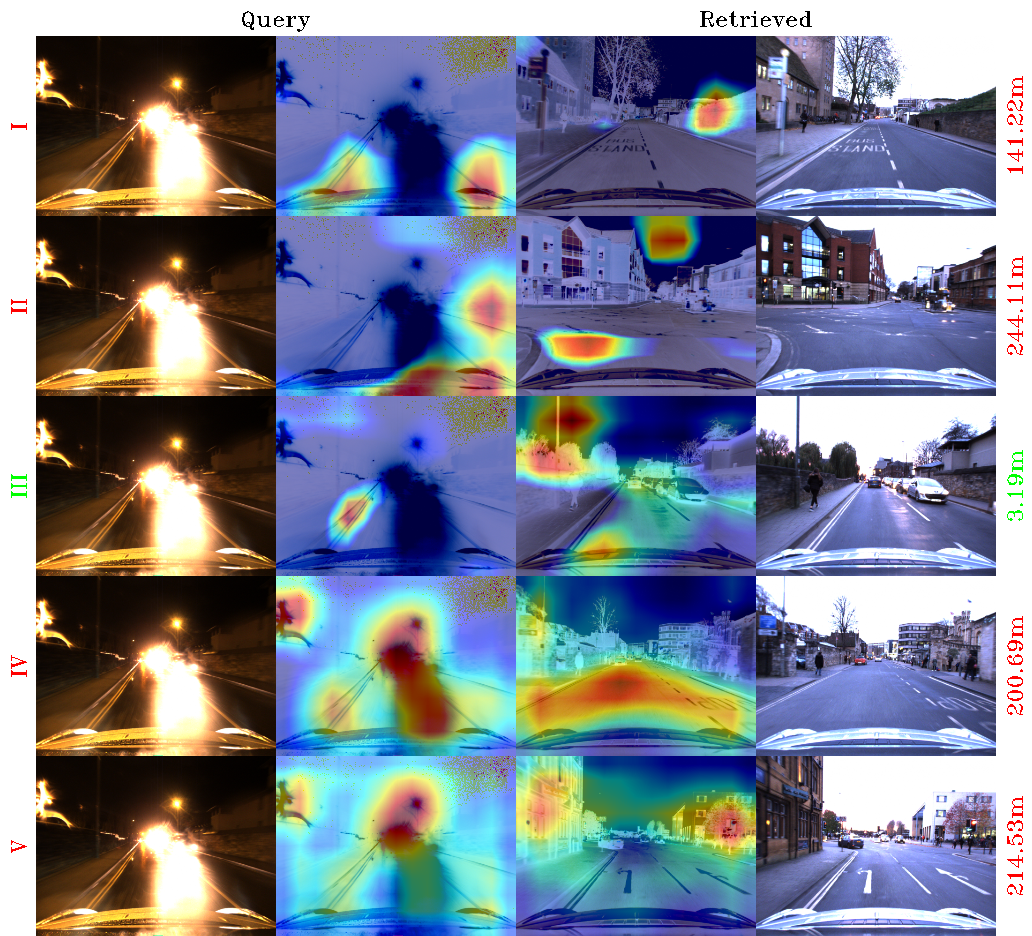}}&
\parbox{0.5\linewidth}{\includegraphics[width=\linewidth]{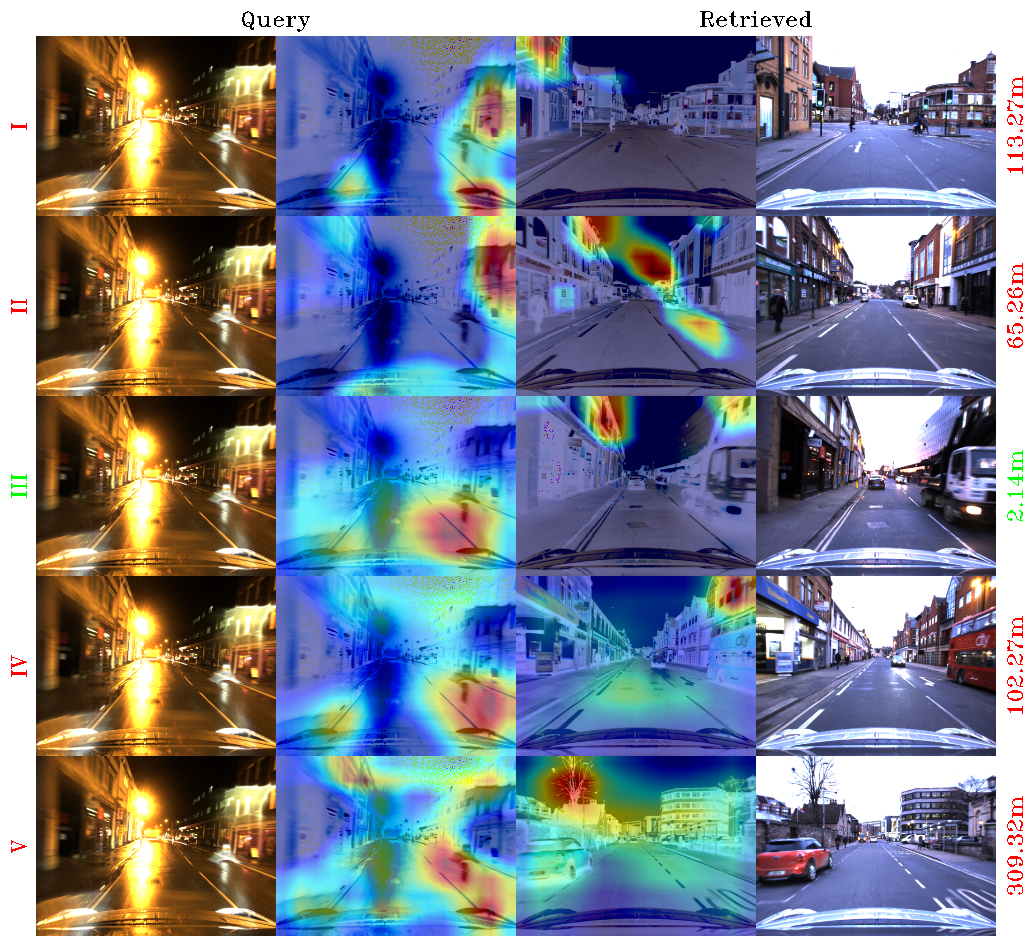}}\\
\\
\toprule
\multicolumn{2}{c}{Cold Freiburg}\\
Sunny & Cloudy\\ \hline
\parbox{0.5\linewidth}{\includegraphics[width=\linewidth]{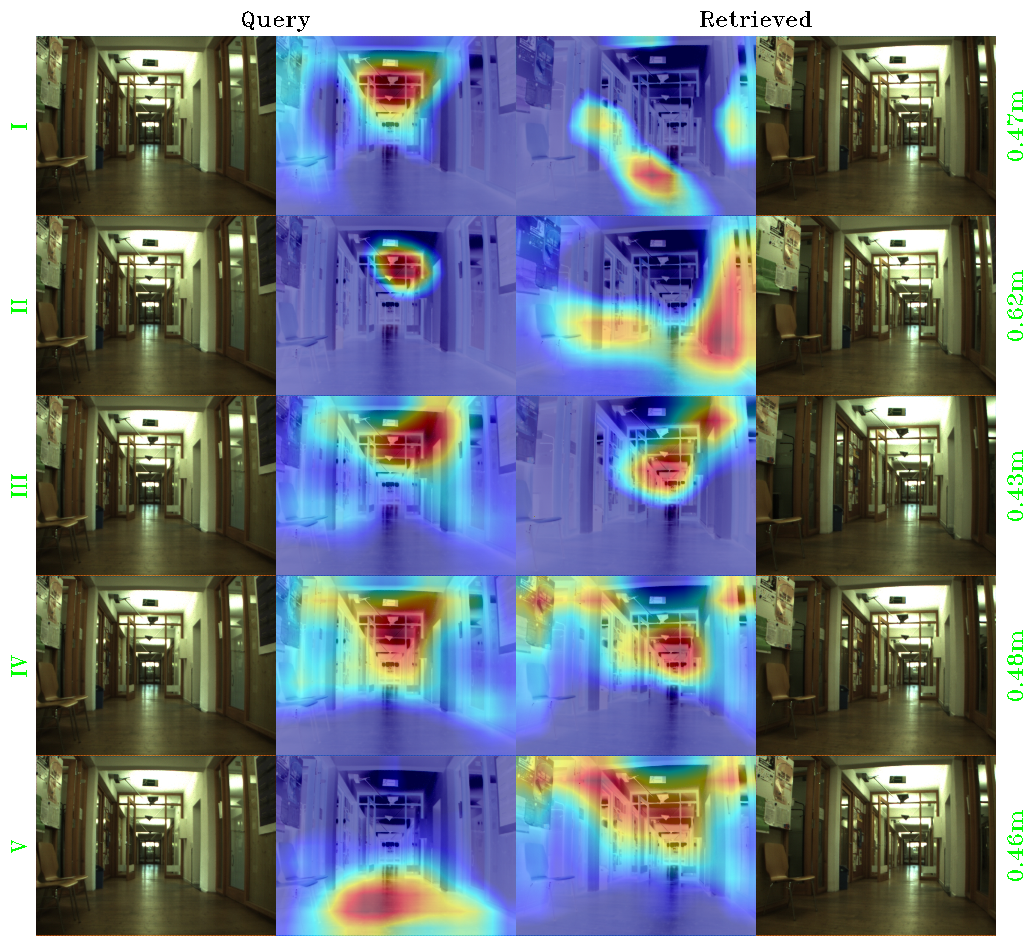}}&
\parbox{0.5\linewidth}{\includegraphics[width=\linewidth]{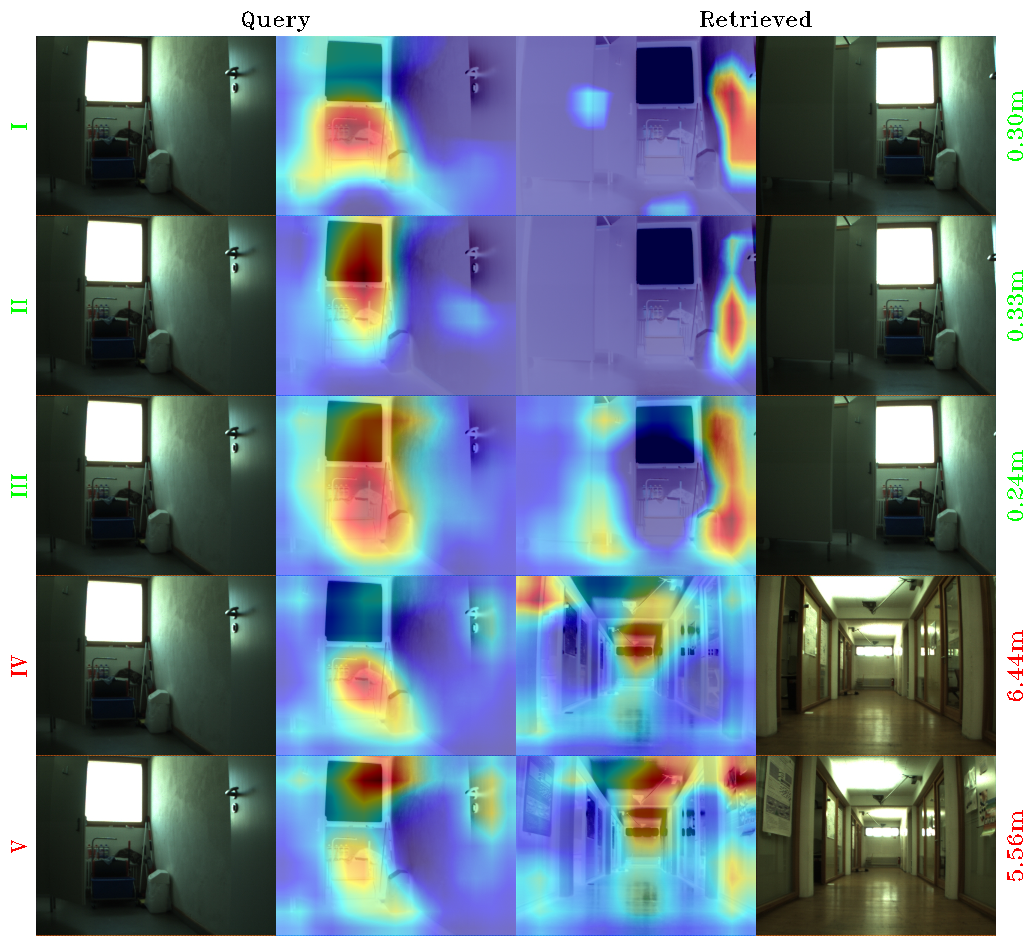}}\\
\\
\toprule
\multicolumn{2}{c}{Pittsburgh}\\ \hline
\parbox{0.5\linewidth}{\includegraphics[width=\linewidth]{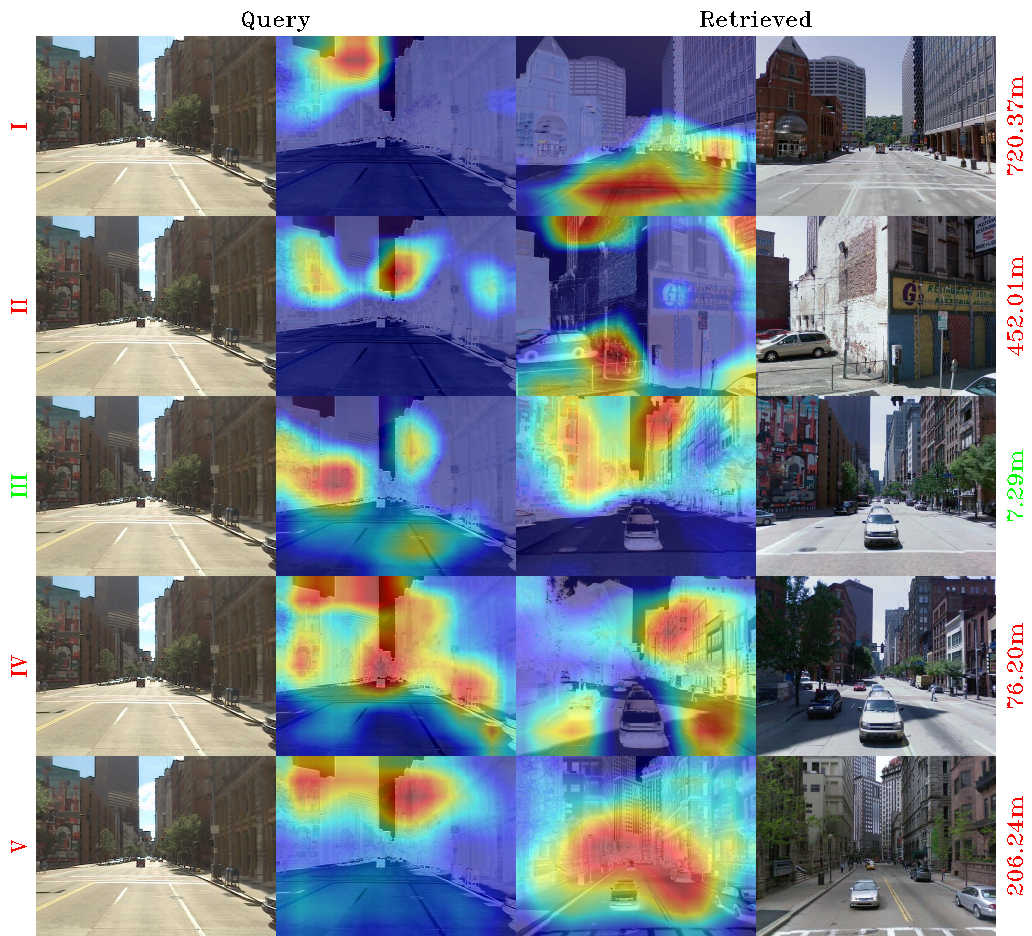}}&
\parbox{0.5\linewidth}{\includegraphics[width=\linewidth]{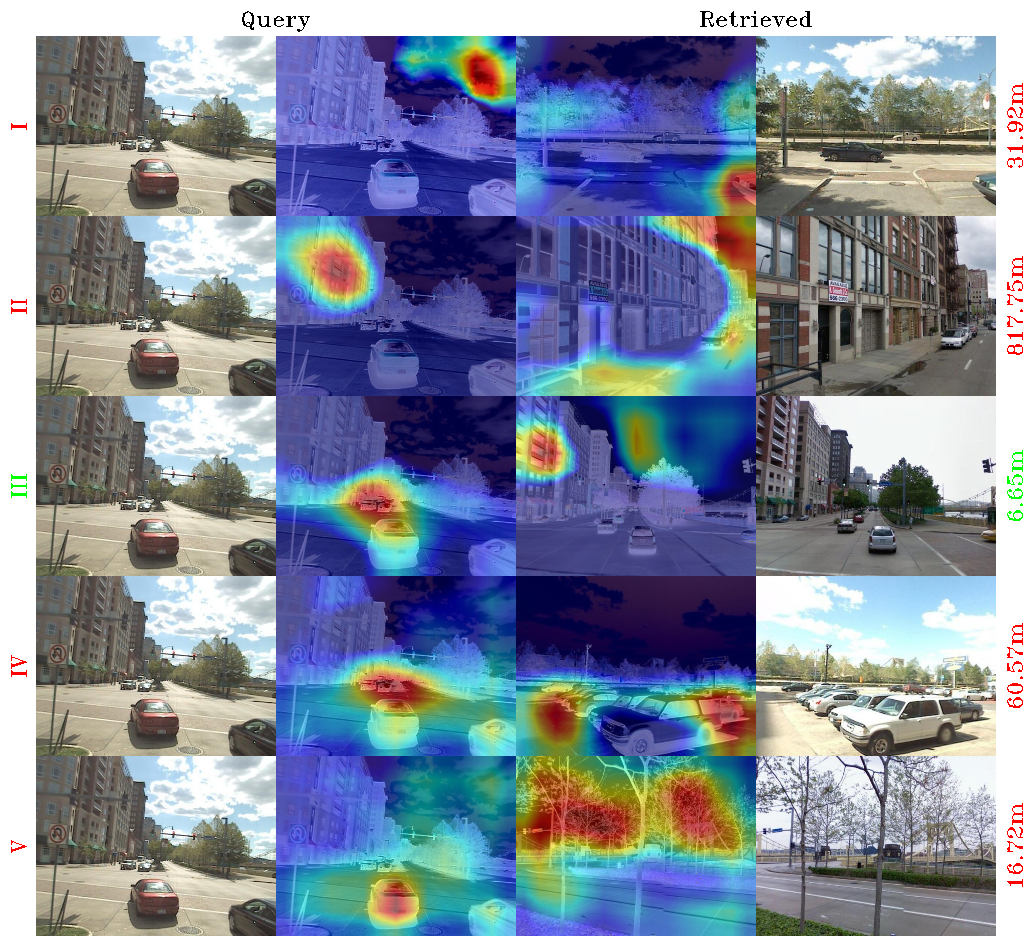}}\\

\end{tabular}
}
\end{center}
\caption{Selected visual results from the best network of each training region---I: Quadruplet on Cold Freiburg, II: Triplet on Oxford RobotCar (small), III: Volume with our mining on Oxford RobotCar (large), IV: Triplet on Pittsburgh30K, V: Initialization with ImageNet, no localization training. For each example we show the query image, Grad-CAM on the query image, Grad-CAM on the retrieved image, retrieved image, and the distance between retrieved image and query image.}
\label{fig:visuals_2}
\end{figure*}

We use Grad-CAM \cite{selvaraju2017grad} to visualize which regions of a query image and the corresponding retrieved reference image contribute most to the match between query and reference image. In other words, we take the gradient of the negative squared feature distance between query image and top-1 retrieved reference image flowing into the last fully convolutional layer (5\_3) to get a coarse heat-map of which regions are most important for the matching decision.
Some selected results are shown in Figure~\ref{fig:visuals_1} and Figure~\ref{fig:visuals_2}.
We also provide a supplementary movie with a comprehensive collection of Grad-CAM visualizations for \emph{III-Ours}: Volume-based with hard positives and pairwise negatives on Oxford RobotCar (large), \emph{IV-Pittsburgh}: Triplet on Pittsburgh30K, \emph{V-Off-the-shelf}: Initialization with ImageNet without localization training.
The query images in the movie are linearly down-sampled to obtain a movie which is roughly five minutes long.
The color of the reported distance between query and reference image indicates whether an image was localized within the specified threshold (10m for outdoors and 1m for Cold Freiburg).

Comparing the results in Figure~\ref{fig:visuals_1}, Figure~\ref{fig:visuals_2}, and the movie with Table~\ref{tab:quantitative} shows that features which quantitatively perform well for localization on most test sets are features which localize based on a large proportion of the image while being less prone to emphasize small parts of the image or confounding objects such as cars or light glares. 
Networks that were trained on small datasets (namely I and II) tend to focus on very specific but seemingly random regions of the images. 
A potential explanation may be over-fitting to particularly salient objects and patterns in their comparatively small training region. 
At the same time, Figure~\ref{fig:visuals_1}, Figure~\ref{fig:visuals_2}, and the movie show how the features trained on the large Oxford RobotCar training region III with volume-based loss and mined positives have learned to focus mainly on trees, buildings and patterns on the street.

\end{document}